\documentclass[journal,compsoc]{IEEEtran}
\usepackage{graphicx}
\usepackage{hyperref}
\usepackage{url}
\usepackage{color}
\usepackage{subfigure}
\usepackage{bbm}
\usepackage{multirow}

\renewcommand{\arraystretch}{1.4}

\begin{document}

\title{Joint Video and Text Parsing for Understanding Events and Answering Queries}

\author{Kewei~Tu,~Meng~Meng,~Mun~Wai~Lee,~Tae~Eun~Choe,~Song-Chun~Zhu
\thanks{K. Tu, M. Meng and S.C. Zhu are with the Department of Statistics, University of California, Los Angeles.}
\thanks{M.W. Lee is with Intelligent Automation, Inc.}%
\thanks{T.E. Choe is with ObjectVideo Inc.}}

\IEEEcompsoctitleabstractindextext{%
\begin{abstract}
We propose a multimedia analysis framework to process video and text jointly for understanding events and answering user queries. Our framework produces a parse graph that represents the compositional structures of spatial information (objects and scenes), temporal information (actions and events) and causal information (causalities between events and fluents) in the video and text. 
The knowledge representation of our framework is based on a spatial-temporal-causal And-Or graph (S/T/C-AOG), which jointly models possible hierarchical compositions of objects, scenes and events as well as their interactions and mutual contexts, and specifies the prior probabilistic distribution of the parse graphs.
We present a probabilistic generative model for joint parsing that captures the relations between the input video/text, their corresponding parse graphs and the joint parse graph. Based on the probabilistic model, we propose a joint parsing system consisting of three modules: video parsing, text parsing and joint inference. Video parsing and text parsing produce two parse graphs from the input video and text respectively. The joint inference module produces a joint parse graph by performing matching, deduction and revision on the video and text parse graphs.
The proposed framework has the following objectives: Firstly, we aim at deep semantic parsing of video and text that goes beyond the traditional bag-of-words approaches; Secondly, we perform parsing and reasoning across the spatial, temporal and causal dimensions based on the joint S/T/C-AOG representation; Thirdly, we show that deep joint parsing facilitates subsequent applications such as generating narrative text descriptions and answering queries in the forms of who, what, when, where and why.
We empirically evaluated our system based on comparison against ground-truth as well as accuracy of query answering and obtained satisfactory results.
\end{abstract}

}

\maketitle

\section{Introduction}
\subsection{Motivation and objective}\label{sec:intro:mot}
Video understanding is an important aspect of multimedia content analysis that aims at automated recognition of the objects, scenes and events in the video. 
It is a crucial component in systems that improve human-computer interaction, such as (a) automatic text generation from videos to facilitate search and retrieval of visual information from the web, and (b) surveillance systems that answer human queries of who, what, when, where and why.
However, understanding contents from video alone has been found to be challenging because of factors such as low resolution, deformation and occlusion; in addition, some aspects of objects and events are hidden, for example, a person's intentions and goals.

\begin{figure}\centering
\includegraphics[scale=0.6]{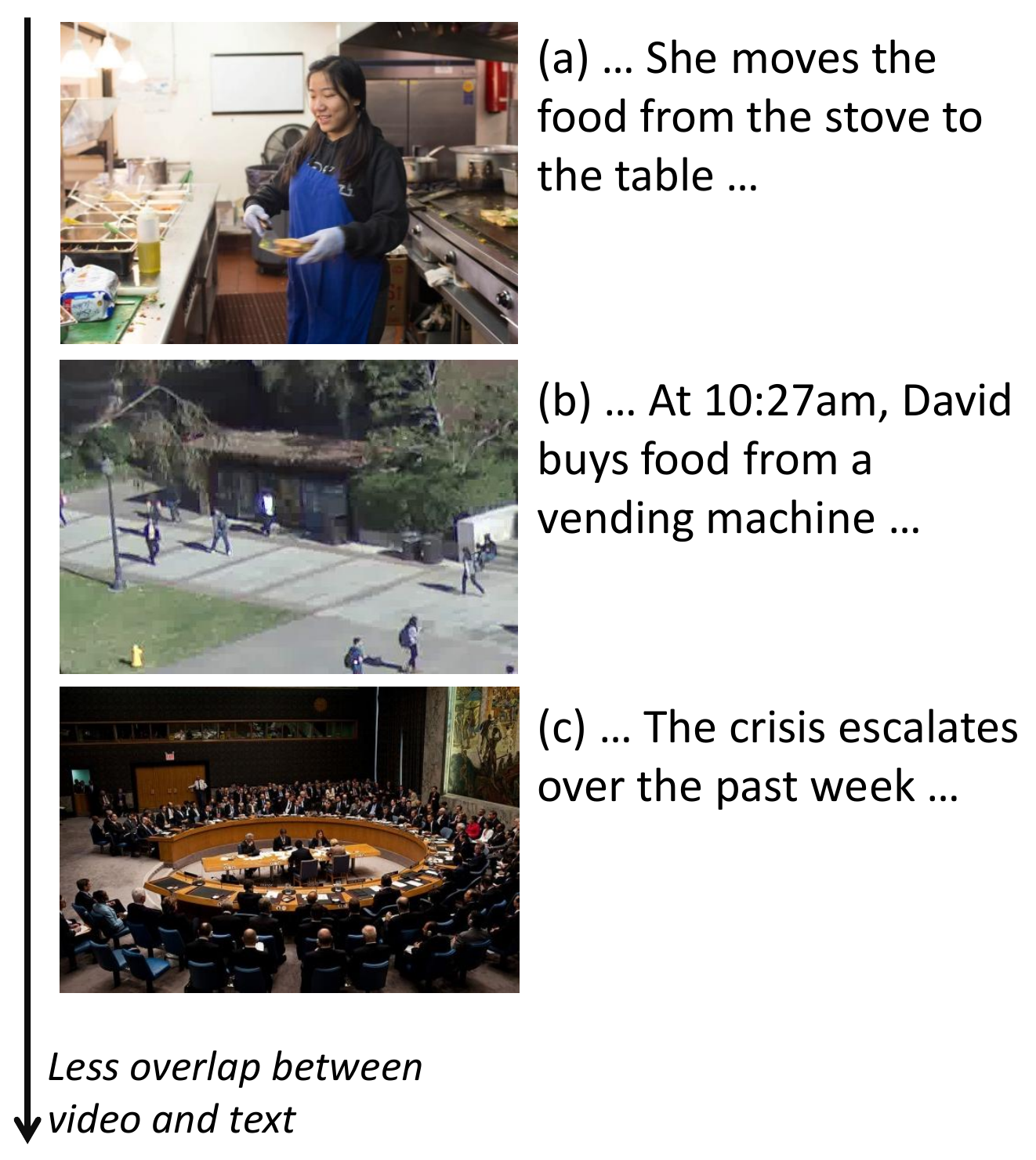}
\caption{Examples of videos accompanied with texts. (a) A television show footage and its screenplay; (b) A surveillance video and the human intelligence descriptions; (c) A news report footage and its closed captions.}\label{fig:examples}
\end{figure}

A significant portion of videos are accompanied by textual descriptions, which provide supplementary information for video understanding. 
Figure \ref{fig:examples} shows three examples, with decreasing levels of overlap between the video contents and the text contents. (a) Movies or television shows and their screenplays, which typically have the largest overlap. (b) Surveillance videos and human intelligence in the form of narrative text descriptions of the scenes and events, which are usually partially overlapped. (c) News report footages and their closed captions of the news anchor's narrative as well as a few lines of on-screen text summarizing the news, which often have the least overlap.

There have been growing interests in multimedia research to process video and text jointly 
and therefore integrate vision and language -- the two major sensory modalities for human intelligence. 
On the one hand, vision provides rich sensory signals to ground language symbols denoting objects, scenes and events; it supplements language with low-level details that are typically not mentioned in text descriptions.
On the other hand, language conveys semantic concepts and relations underlying the visual world, some of which may be hidden or obscure from the visual channel (such as names, certain attributes, functions, intents, and causality).
Furthermore, since vision and language provide information of the same scene from two different channels, they could disambiguate and reinforce each other and produce a more accurate joint interpretation.

In contrast to the bag-of-words representation widely used in the video understanding and multimedia literature (e.g., \cite{Feng04mb,Barnard03mw,MonayG07ms,Wang09si}), in this work we propose to represent the joint interpretation of video and text in the form of a parse graph.
A parse graph is a labeled directed graph that can be seen as an extension of the constituency-based parse tree used in natural language syntactic parsing \cite{Manning99book}. 
It has been employed in computer vision to model objects \cite{Zhu06as}, scenes \cite{Zhao11ip,Zhao13sp} and events \cite{Pei11pv,Fire13uc}. 
A node in a parse graph represents an entity that can be an object, an event or a status of an object (i.e., fluent). An edge in a parse graph represents a relation between two entities. Together they represent the compositional structures of spatial information (for objects and scenes), temporal information (for actions and events) and causal information (causalities between events and object statuses).
Such an explicit semantic representation facilitates subsequent applications such as text generation and query answering.

In order to handle different levels of overlap in the video and text, joint parsing must overcome the following challenges.
\begin{itemize}
\item When overlap exists between video and text, we need to solve the coreference problem, i.e., identifying the correspondence between the entities and relations observed in the video and those described in the text. For example, we may need to identify that a person detected in the video and a name mentioned in the text correspond to the same person. This problem can be complicated when multiple entities or relations of the same type appear in the same scene.
\item When there is little overlap between video and text, we need to fill in additional information to construct a coherent and comprehensive joint interpretation. For example, if the video shows a person waiting at a food truck and the text mentions a person buying food, then we need to infer that waiting at a food truck is typically followed by buying food in order to connect the video and text.
\item Occasionally, the content detected from the video may conflict with the content of the text, because of either erroneous parsing or inaccurate input. We need to resolve the conflicts and provide a consistent joint interpretation.
\end{itemize}

\subsection{Overview of our work}
\begin{figure}\centering
  \includegraphics[scale=0.4]{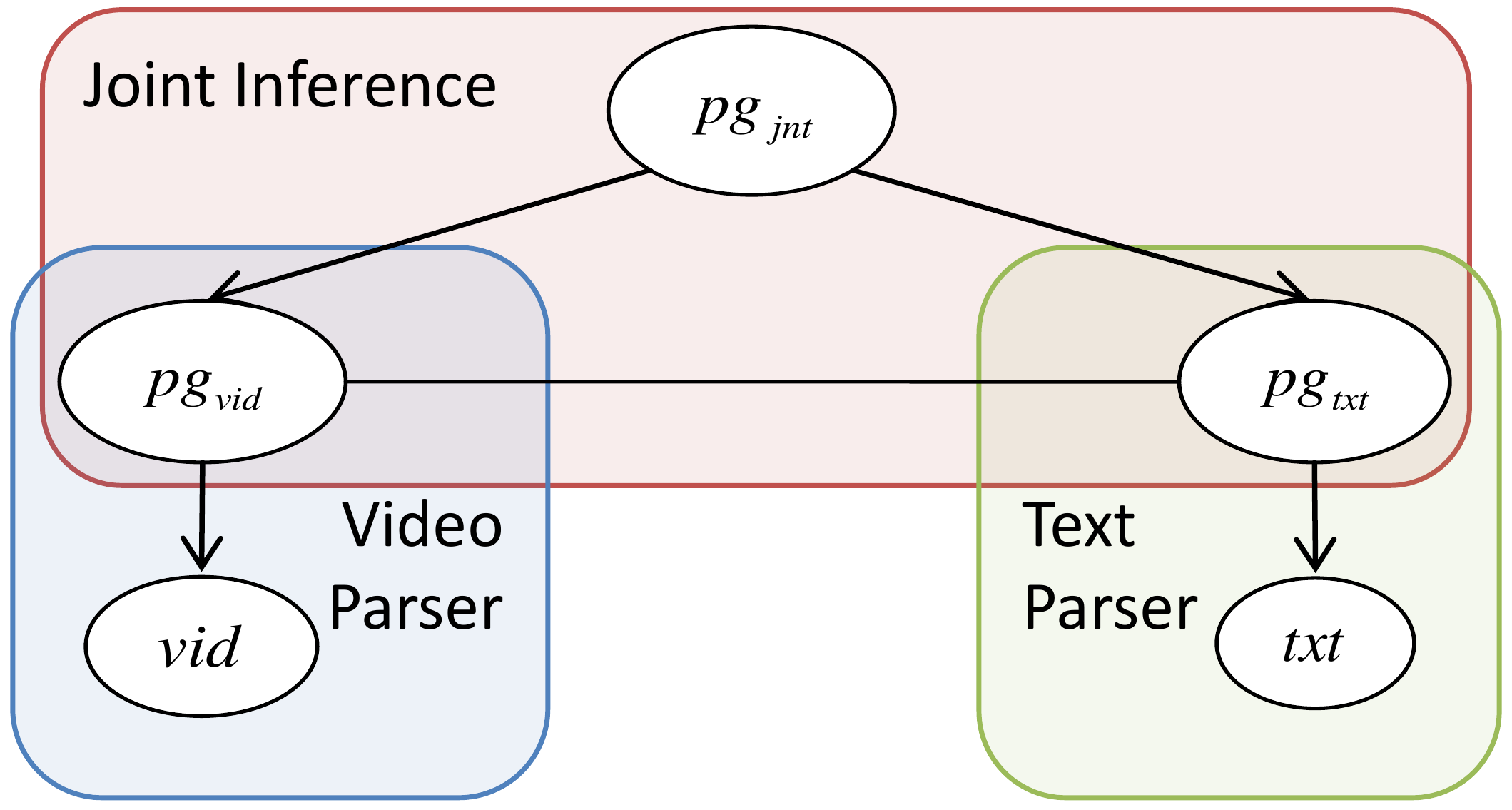}\\
  \caption{The graphical model for joint parsing of video and text. See text for details.}\label{fig:model}
\end{figure}

We focus on joint parsing of surveillance videos and narrative text descriptions, but our framework can be employed or extended to process other types of video-text data as well.
Figure \ref{fig:model} shows our probabilistic generative model. Let $vid$ and $txt$ denote the input video and text respectively, $pg_{vid}$ the parse graph that interprets the input video, $pg_{txt}$ the parse graph that interprets the input text, and $pg_{jnt}$ the joint parse graph for both the video and the text.
Our goal can be formulated as optimizing the posterior probability of the three parse graphs $pg_{jnt}, pg_{vid}$ and $pg_{txt}$ given the input video $vid$ and text $txt$.
\begin{eqnarray}\label{eq:obj}
    \lefteqn{ P(pg_{jnt}, pg_{vid}, pg_{txt}|vid,txt) } \\
    & \propto & P(pg_{jnt}) P(pg_{vid},pg_{txt}|pg_{jnt}) P(vid|pg_{vid}) P(txt|pg_{txt}) \nonumber
\end{eqnarray}

Our system for optimizing this posterior probability can be divided into three modules.
First, a \emph{video parser} generating candidate video parse graphs $pg_{vid}$ from the input video $vid$. The video parser encodes the conditional probability $P(vid|pg_{vid})$. It is also guided by a spatial-temporal-causal And-Or graph (S/T/C-AOG) \cite{Zhu06as,Zhao11ip,Pei11pv,Fire13uc}, which models the compositional structures of objects, scenes and events.
Second, a \emph{text semantic parser} parsing the text descriptions $txt$ into the text parse graph $pg_{txt}$. We build our semantic parser on top of the Stanford Lexicalized Parser \cite{SP}, focusing on parsing narrative event descriptions to extract information about objects, events, fluent changes and the relations between these entities. The text parser specifies the conditional probability $P(txt|pg_{txt})$.
Third, a \emph{joint inference module} producing a joint parse graph $pg_{jnt}$ based on the video and text parse graphs produced by the first two modules. The joint inference takes into account 1) the relationships between the joint parse graph and the video and text parse graphs, which are encoded in the conditional probability $P(pg_{vid}, pg_{txt} | pg_{jnt})$, and 2) the compatibility of the joint parse graph with the background knowledge represented in the S/T/C-AOG, as encoded in the prior probability $P(pg_{jnt})$. We employ three types of operators in constructing the joint parse graph:
\begin{itemize}
\item Solving the coreference problem by \emph{matching} the video and text parse graphs
\item Filling in the potential gap between video and text by \emph{deduction}.
\item Resolving possible conflicts via \emph{revisions} to the video and text parse graphs.
\end{itemize}

\begin{figure*}\centering
  \includegraphics[scale=0.55]{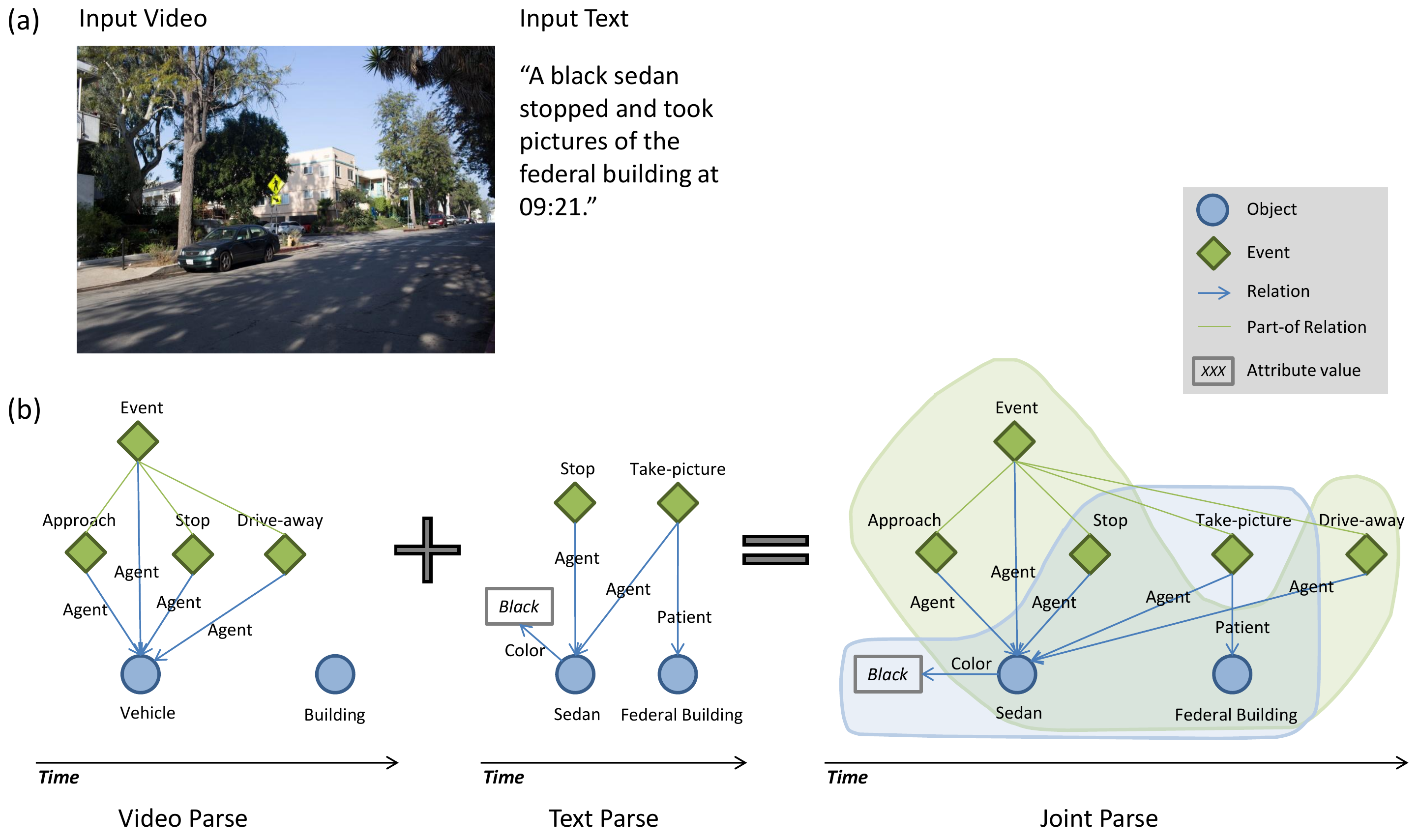}\\
  \caption{(a) An example surveillance video and the text descriptions of a parking lot scene. (b) The video parse graph, text parse graph and joint parse graph. The video and text parses have significant overlap. In the joint parse graph, the green shadow denotes the subgraph from the video and the blue shadow denotes the subgraph from the text.}\label{fig:suv_ex}
\end{figure*}

\begin{figure*}\centering
  \includegraphics[scale=0.55]{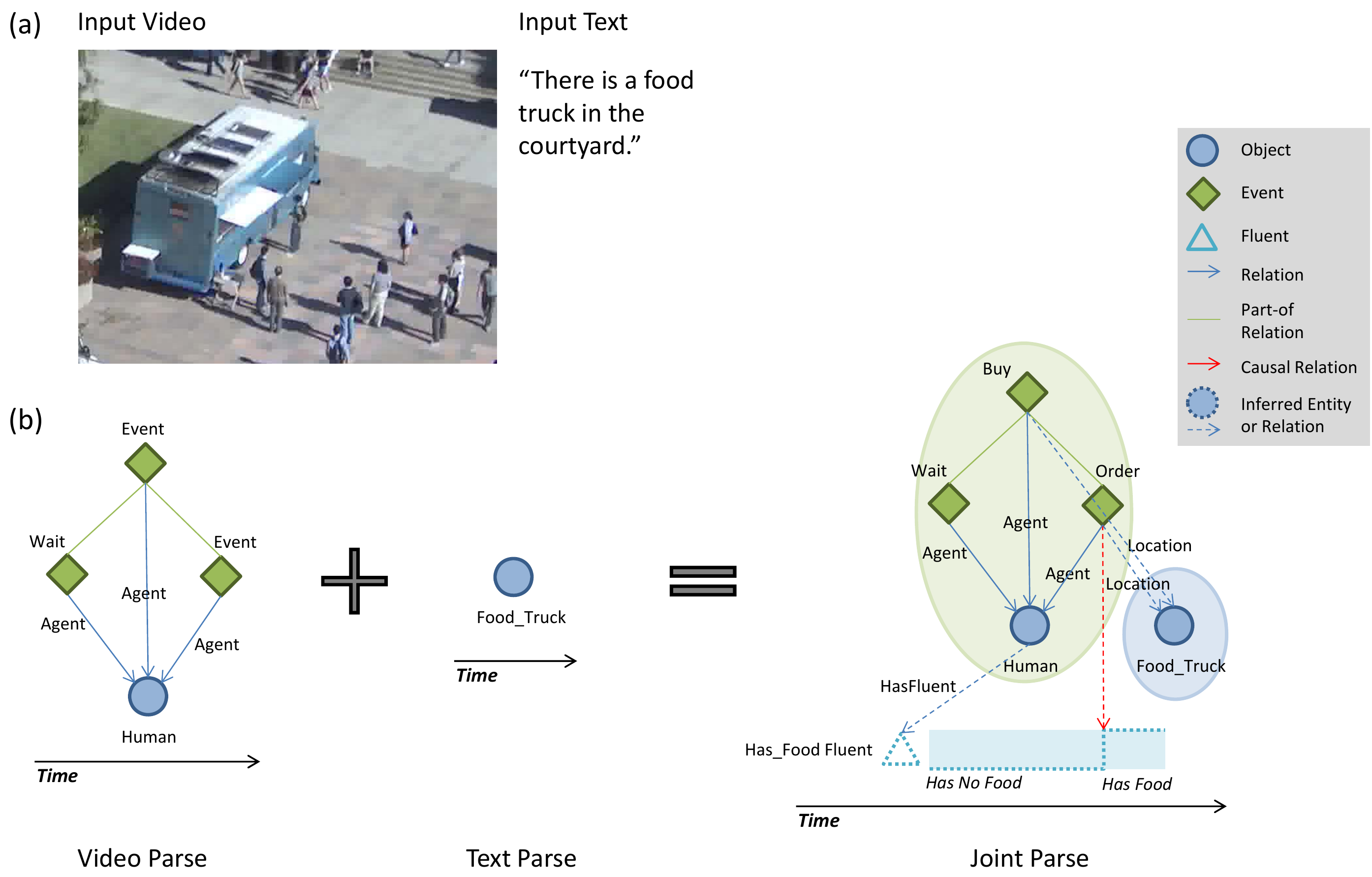}\\
  \caption{(a) An example surveillance video and the text descriptions of a courtyard scene. (b) The video parse graph, text parse graph and joint parse graph. The video and text parses have no overlap. In the joint parse graph, the green shadow denotes the subgraph from the video and the blue shadow denotes the subgraph from the text.}\label{fig:foodtruck_ex}
\end{figure*}

Figure \ref{fig:suv_ex} and \ref{fig:foodtruck_ex} show two examples of joint parsing from videos and texts with different levels of overlap.
In each parse graph shown in the figures, we use different shapes to represent objects, events and fluents, and use different edge colors to represent different relations; the label on a node or edge indicates the semantic type of the denoted entity or relation. For each event node, the \texttt{Agent} relation connects it to the action initiator, and the \texttt{Patient} relation connects it to the action target. The set of event nodes are laid out horizontally based on their occurring time and vertically based on their compositional complexity. For a fluent node, we use levels to indicate its value at each time point.
Figure \ref{fig:suv_ex}(a) shows a surveillance video and the human intelligence description of a road scene, in which the video and text cover the same event and have significant overlap. Figure \ref{fig:suv_ex}(b) shows the results of video parsing, text parsing and joint inference. The video and text parses reinforce each other in the joint parse via overlapping (e.g., the stopping event, the building), and they also supplement each other by providing additional information (e.g., the taking-picture event is too subtle to recognize from the video but is provided in the text).
Figure \ref{fig:foodtruck_ex}(a) shows a surveillance video and text description of a courtyard scene. The video shows a few people waiting and taking turns, but their target is not recognized by the video parser; the text only mentions a food truck. The video and text parses have no direct overlap in terms of object entities, but these objects are involved in a common food purchase event. Figure \ref{fig:foodtruck_ex}(b) shows the results of video parsing, text parsing and joint inference. The joint parse graph identifies the food truck as the target of people's actions and infers the types of the events as well as the fluent changes resulting from the events, thus providing a more coherent and comprehensive interpretation of the scene.

We encode the prior knowledge of parse graphs in the S/T/C-AOG and use it to guide both video parsing and joint inference, which is another contribution of this work. An AOG is an extension of a constituency grammar used in natural language parsing \cite{Manning99book} to represent hierarchical compositions of objects, scenes and events.
An AOG has a hierarchical structure with alternating layers of And-nodes and Or-nodes. An And-node represents the configuration of a set of sub-entities to form a composite entity (e.g., waiting followed by ordering composes a buying event). An Or-node represents the set of alternative compositional configurations of an entity (e.g., a buying event can be either ``waiting and ordering'' or ``ordering without waiting'').
\begin{itemize}
\item A spatial And-Or graph (S-AOG) \cite{Zhu06as,Zhao11ip} models the spatial decompositions of objects and scenes.
\item A temporal And-Or graph (T-AOG) \cite{Pei11pv} models the temporal decompositions of events to sub-events and atomic actions.
\item A causal And-Or graph (C-AOG) \cite{Fire13uc} models the causal decompositions of events and fluent changes.
\end{itemize}
These three types of AOGs are interconnected and together they form a joint S/T/C-AOG.
A parse graph is an instance of an AOG that selects only one of the alternative configurations at each Or-node of the AOG. The prior probability of a parse graph is specified by the energy of configuration selections made at the Or-nodes of the AOG as well as the energy of the compositional configurations at the And-nodes of the AOG contained in the parse graph.

\subsection{Evaluation}
To empirically evaluate our system, we collected two datasets, each containing surveillance videos of daily activities in indoor or outdoor scenes and a set of text descriptions provided by five human subjects. We asked each human subject to provide text descriptions at three levels of details, so we can study how the amount of text descriptions and their degrees of overlap with the video content can impact the performance of joint parsing. The joint parse graphs produced by our system were then evaluated using two different approaches.

In the first approach, we compare the joint parse graphs against the ground truth parse graphs which were constructed based on the merged parses of all the text descriptions from all the human subjects. We measure the precision and recall where precision is the percentage of the joint parse graph that is correct (i.e., contained in the ground truth) and recall is the percentage of the ground truth that is covered by the joint parse graph.

One important application of video understanding is to answer queries from human users. Therefore, in the second evaluation approach, we measure the accuracy of a query answering system based on the joint parse graph produced by our system. We collected a set of natural language queries that mimics the natural human-computer interaction and includes questions in the forms of who, what, when, where and why. The answers produced by the system were then compared against the correct answers provided by human subjects.

The evaluation results show that the joint parse graphs provide more accurate and comprehensive interpretations of the scenes than the video and text parse graphs. In addition, we find that additional text input improves the joint parse graphs but with diminishing return.

\subsection{Related work}
\subsubsection{Bag-of-words based approaches}
There has been a large body of work in the areas of multimedia, computer vision and machine learning on joint processing of video/image and text. The majority of the existing work is based on the bag-of-words (BoW) representation for both video/image and text. In the BoW representation, a text is represented as a set of unordered words; and a video/image is represented as a set of unordered visual words, each of which is a feature vector (e.g., SIFT and HoG) of an image patch. 
The BoW representation is easy to extract and process, but it ignores the compositional structures based on the words. Therefore, BoW-based approaches typically perform coarse-level processing of the video/image and text, e.g., classification or retrieval, instead of producing a full interpretation of the input video/image and text.
For example, for joint video and text processing,
Feng et al. \cite{Feng04mb} applied a joint model of video/image and text for both automatic image annotation and retrieval;
Iyengar et al. \cite{Iyengar05jv} modeled videos and text jointly for the purpose of multimedia information retrieval;
Yang et al. \cite{Yang07hm} addressed the problem of video classification by joint modeling of videos and transcript texts;
Xu et al. \cite{Xu08uw} performed sports video semantic event detection based on joint analysis of videos and webcast texts.
For joint processing of image and text, 
Paek et al. \cite{Paek1999iv} performed scene classification with a TF-IDF based approach applied to both images and the accompanying text;
Barnard et al. \cite{Barnard03mw} and Blei \cite{Blei04pm} discussed a series of probabilistic models to capture the dependencies between image elements and words;
Berg and Forsyth \cite{Berg2006aw} demonstrated a method for identifying images containing categories of animals based on word and visual cues;
Monay and Gatica-Perez \cite{MonayG07ms} applied the probabilistic latent semantic analysis model for annotated images to improve automatic image indexing;
Wang et al. \cite{Wang09si} simultaneously modeled the image elements, text annotations and image class labels using a probabilistic topic model;
Liu et al. \cite{Liu2011tq} presented an image retrieval system by leveraging large-scale web image and their associated textual descriptions;
Jia et al. \cite{Jia11lc} proposed a probabilistic model that encodes relations between loosely related images and text descriptions;
Feng and Lapata \cite{Feng13ac} learned a probabilistic topic model of visual and textual words for the purpose of automatic image caption generation.
In contrast to the existing work, our work aims at deep semantic parsing of video and text, i.e., we try to identify the compositional structures contained in the video and text, based on which we extract and explicitly represent a full interpretation of the input video and text. Our scope of parsing includes spatial parsing (of objects and scenes), temporal parsing (of actions and events) and causal parsing (of events and fluents), some of which have rarely been explored in the previous work of joint processing of video/image and text.

\subsubsection{Parsing based approaches}
Our work is built on the recent advancement on image/video parsing, which identifies the hierarchical compositional structures contained in the image or video. In particular,
spatial parsing identifies the spatial compositional structures of objects and scenes \cite{Tu05ip,Zhu06as,Jin06ca,Fidler07ts,Han09bt,Porway10ah,Zhao11ip,Zhao13sp}; temporal parsing recognizes the temporal compositional structures of events \cite{Brand97ch,Natarajan07ch,Gorga10cs,Ryoo06ro,Joo06ro,Zhang11ae,Pei11pv,Amer2012cs}; and causal parsing detects the causal relations between events and fluent changes \cite{Hakeem04ca,Brendel11pe,Fire13uc}.


Our work also involves semantic parsing of text descriptions, which is related to a variety of semantic parsing approaches studied in the natural language processing community, for example, a supervised approach \cite{Mooney09lf}, an unsupervised approach \cite{Poon09us}, and a dependency based approach \cite{Liang11ld}. Semantic text parsing is typically built on top of a syntactic parser such as the Stanford Lexicalized Parser \cite{SP}. 

For joint parsing of video and text, Hobbs and Mulkar-Mehta \cite{Hobbs08ua} described a preliminary framework that interprets the combination of video and text from news broadcasts, which is the work most closely related to ours. Compared with their work, our system extracts information directly from the raw data of video and text, and relies on a probabilistic generative model to handle uncertainty; besides, we have done comprehensive experiments to evaluate our system.
The graph-based referential grounding approach proposed by Liu et al. \cite{Liu2012tm} is also related to our work, but they focused on spatial relations between objects while we study spatial-temporal-causal parsing of objects, scenes and events.

\begin{figure*}\centering
  \includegraphics[width=\textwidth]{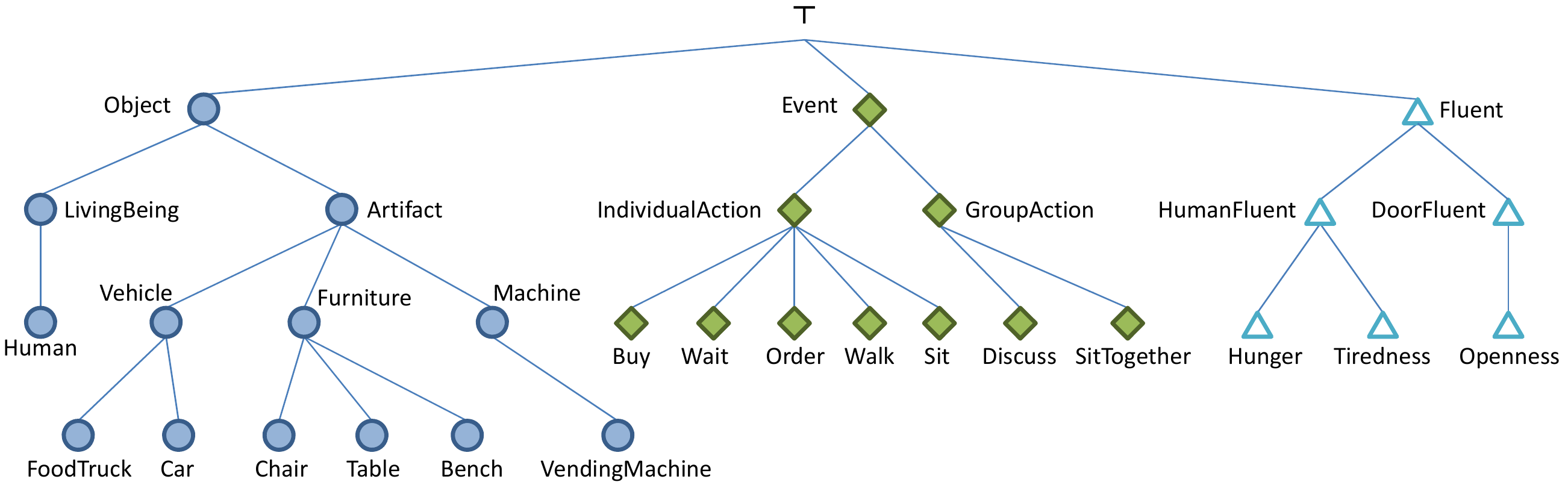}\\
  \caption{The taxonomy in an example ontology for the courtyard scene. The symbol $\top$ at the root node represents the \emph{universal} type, which is a supertype of every other type.}\label{fig:ontology}
\end{figure*}

\subsection{Contributions}
The contributions of our paper include three aspects:
\begin{itemize}
\item Although there has been a lot of previous work in video parsing and text semantic parsing, our work is the first to jointly parse video and text for more accurate and comprehensive parsing results. Our work of joint semantic parsing also goes beyond the traditional bag-of-words approaches to joint processing of video and text.
\item Although spatial, temporal and causal AOGs and their parsing techniques have been separately investigated before, our work is the first to propose the joint S/T/C-AOG which crosses the spatial, temporal and causal dimensions and enables the propagation of information between them during parsing.
\item We propose novel evaluation approaches of video understanding based on parse graph comparison and query answering through storylines. In particular, the query-answering-based evaluation mimics the natural human-computer interaction via questions of who, what, when, where and why, which goes beyond the conventional evaluation frameworks such as those based on classification or detection.
\end{itemize}

\subsection{Outline of the paper}
The rest of the paper is organized as follows. Section \ref{sec:rep} introduces our knowledge representations. Section \ref{sec:video} and \ref{sec:text} present the approaches of video parsing and text parsing respectively. Section \ref{sec:joint} introduces our joint inference module. Section \ref{sec:query} introduces the application of joint parsing in natural language query answering. Section \ref{sec:exp} presents our experiments. Section \ref{sec:conc} concludes the paper.

\section{Representations}\label{sec:rep}
\subsection{Ontology}\label{sec:rep:onto}
An ontology is a ``formal, explicit specification of a shared conceptualisation'' \cite{Gruber93at}. We use an ontology to specify the types of entities and relations that may appear in the parse graphs and therefore define the scope of our study. This ontology can be manually constructed using an ontology editor (e.g., \cite{Gennari03te,Corcho02wa,Zhang04oi}), adapted from an existing ontology (e.g., \cite{WordNet,Cyc}), or automatically or semi-automatically learned from data (e.g., \cite{Maedche01ol,Buitelaar05ol,Cimiano06ol}).

As typical for an ontology, we organize the types of entities (i.e., concepts) into a taxonomy. A taxonomy is a directed acyclic graph in which a directed edge between two concepts denotes a supertype-subtype relationship. As will be shown in section \ref{sec:joint}, we rely on the taxonomy to compute semantic distances between concepts when doing joint inference. We divide the concepts into three main categories at the top level of the taxonomy.
\begin{itemize}
\item Object, which represents any physical objects including human.
\item Event, which includes any basic actions (e.g., sitting and walking) as well as any activities that are composed of a sequence of actions (e.g., buying). 
\item Fluent, which is used in the AI community to refer to an attribute of a physical object that can change over time \cite{Mueller06cr}, e.g., the openness attribute of a door and the hunger attribute of a human. The change of fluents helps detect or explain human actions in a scene and thus is important in joint parsing.
\end{itemize}
Figure \ref{fig:ontology} shows an example taxonomy for the courtyard scene.

We also specify a set of relations in the ontology. These relations can be organized into a relation taxonomy, allowing the computation of semantic distances between relations. A small set of core relations are introduced in the next section.

\begin{figure*}
\centering
\includegraphics[scale=0.6]{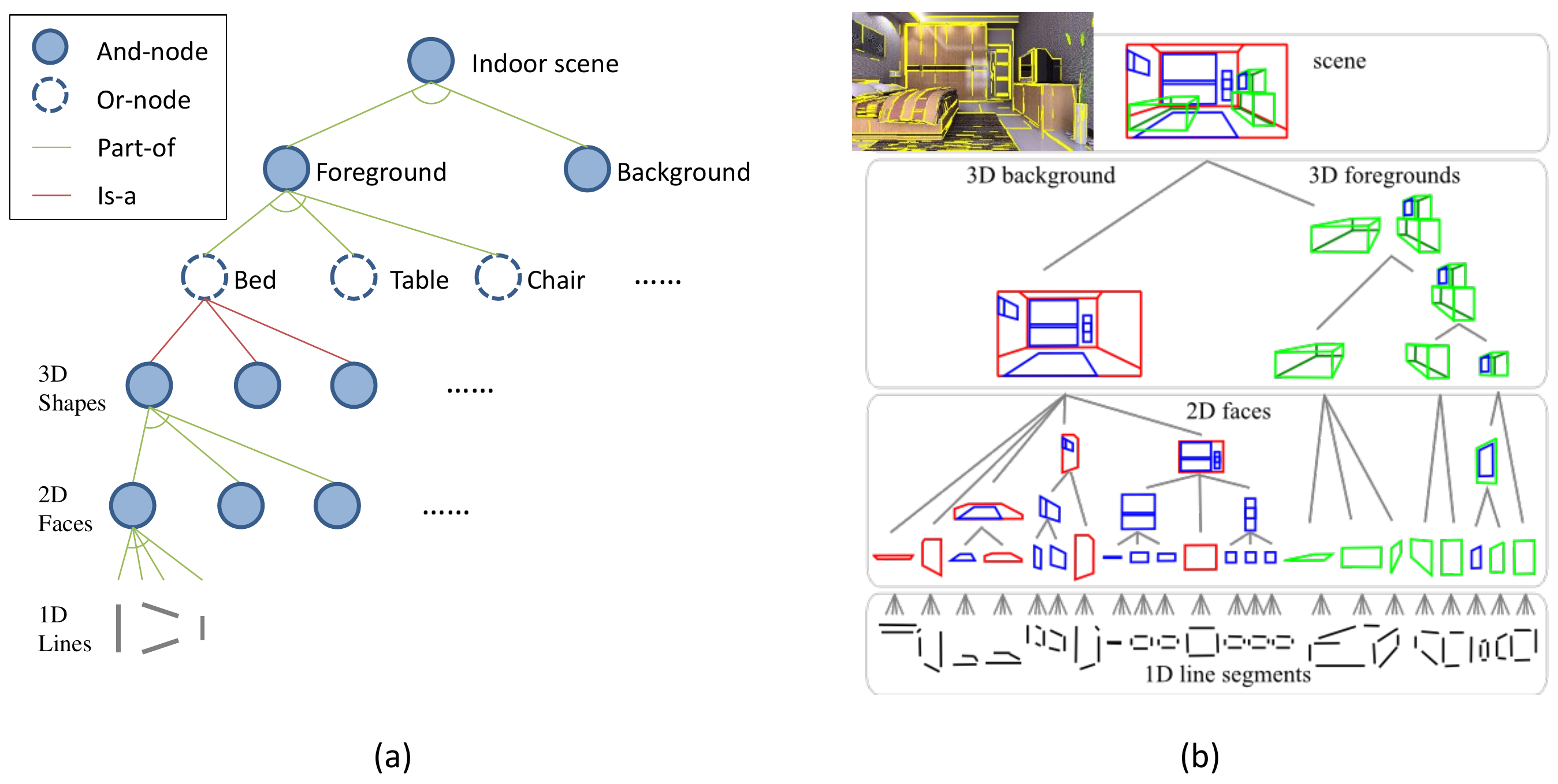}
\caption{(a) An example spatial And-Or graph (S-AOG). Each node represents a type of scenes, objects or their parts; an And-node represents the spatial composition of its child nodes; an Or-node represents a selection among its child nodes; a leaf node represents an atomic object.
(b) An example spatial parse graph from \cite{Zhao11ip}, which is a realization of the S-AOG by making selections at Or-nodes.}
\label{fig:s-aog}
\end{figure*}

\subsection{Knowledge representation by And-Or graphs}\label{sec:rep:aog}
We represent the compositional structures of objects, scenes and events using a spatial-temporal-causal And-Or graph (S/T/C-AOG). We will define probabilistic models on the AOG to address the stochasticity of the compositional structures in section \ref{sec:rep:prob}.
An AOG is an extension of a constituency grammar used in natural language parsing \cite{Manning99book}.
Like a constituency grammar, an AOG represents possible hierarchical compositions of a set of entities. However, an AOG differs from a constituency grammar in that it contains additional annotations and relations over the entities.
More specifically, an AOG consists of a hierarchical structure with alternating layers of And-nodes and Or-nodes. An And-node represents the decomposition of an entity of a specific type into a set of sub-entities. A \texttt{Part-of} relation is established between each sub-entity and the composite entity. The And-node also specifies a set of relations between the sub-entities, which configure how these sub-entities form the composite entity. An Or-node represents the alternative configurations of an entity of a specific type. 
AOGs has been employed in computer vision to model the hierarchical decompositions of objects \cite{Zhu06as}, scenes \cite{Zhao11ip,Zhao13sp} and events \cite{Pei11pv,Fire13uc}. In this work, we use an AOG to model objects, scenes, events and perceptual causal effects in the joint interpretation of video and text.

\subsubsection{Spatial And-Or graph}
A spatial And-Or graph (S-AOG) \cite{Zhu06as,Zhao11ip} models the spatial decompositions of objects and scenes. Figure \ref{fig:s-aog}(a) shows an example S-AOG for indoor scenes.
Each node in the S-AOG represents a type of scenes, objects or their parts.
An And-node in the S-AOG represents a scene or object that is the spatial composition of a set of parts. For example, a table can be the spatial composition of a table top above four legs; and the background of an indoor scene can be the spatial composition of multiple 2D planes (i.e., the walls, floor and ceiling) that are hinged.
An Or-node in the S-AOG represents alternative configurations of a scene or object. For example, a table can have many different styles; and an indoor scene may have a few different typical viewpoints.
A leaf node in the S-AOG represents an atomic object that cannot be further decomposed.

\subsubsection{Temporal And-Or graph}
\begin{figure*}
\centering
\includegraphics[scale=0.6]{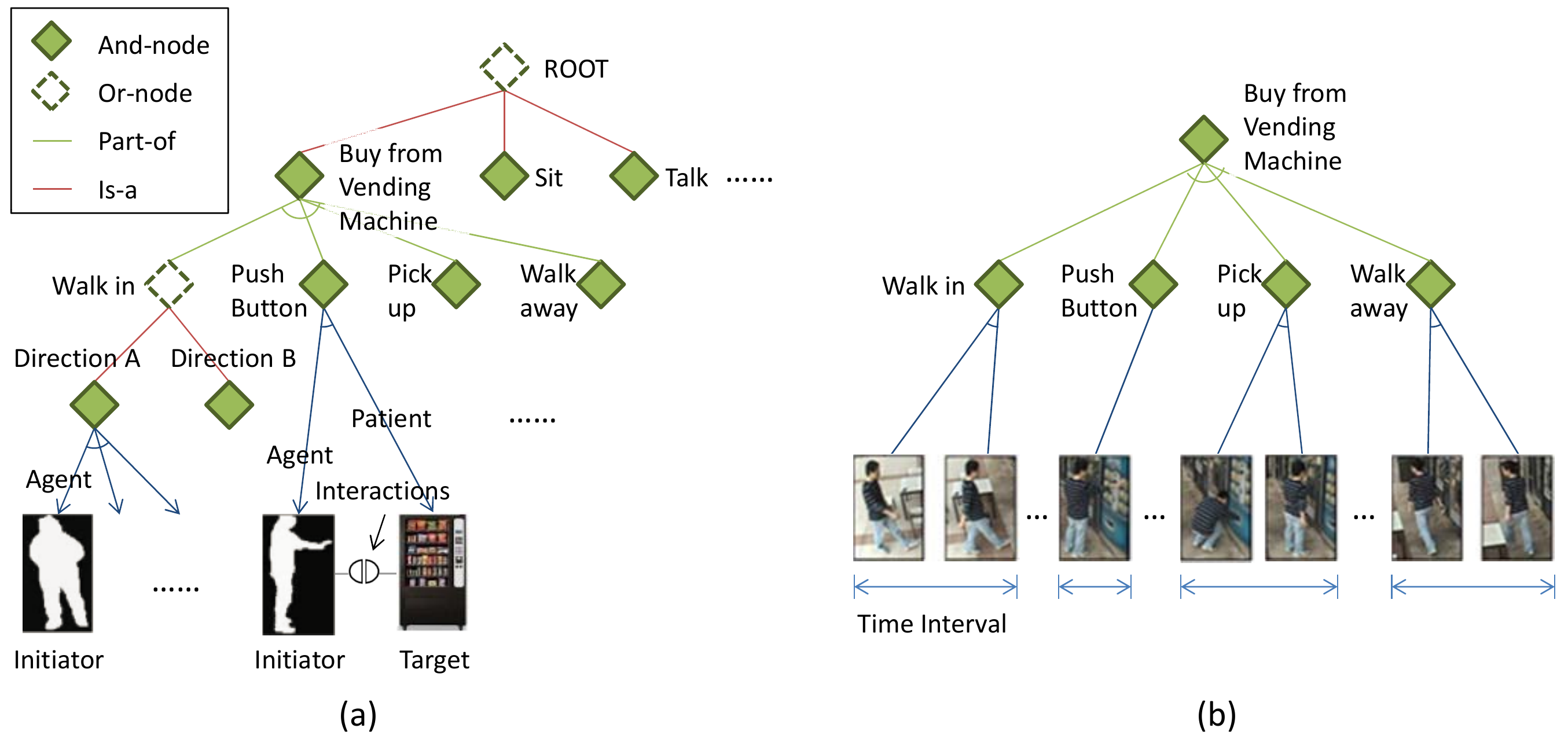}
\caption{(a) An example temporal And-Or graph (T-AOG). Each node represents a type of event or action; an And-node represents the temporal composition of its child nodes; an Or-node represents a selection among its child nodes; a leaf node represents an atomic action defined by the properties of and interactions between the action initiator(s) and target(s).
(b) An example temporal parse graph, which is a realization of the T-AOG by making selections at Or-nodes.}
\label{fig:t-aog}
\end{figure*}

A temporal And-Or graph (T-AOG) \cite{Pei11pv} models the hierarchical decompositions from events to sub-events and then to atomic actions. Figure \ref{fig:t-aog}(a) shows an example T-AOG. 
Each node in the T-AOG represents a type of event or action occurred in a time interval. 
An And-node in the T-AOG represents an event that is the temporal composition of a set of sub-events. For example, a buying event can be the temporal composition of waiting followed by ordering. 
An Or-node in the T-AOG represents alternative configurations of an event, e.g., a buying event can be either ``waiting and ordering'' or ``ordering without waiting''. 
A leaf node in the T-AOG represents an atomic action which is defined by properties of the action initiator(s) and the action target(s) (e.g., the pose of a human, the state of an object) as well as the interactions between the initiator(s) and target(s). We use \texttt{Agent} to denote the relation between an action and the action initiator, and use \texttt{Patient} to denote the relation between an action and the action target.

\subsubsection{Causal And-Or graph}
\begin{figure*}
\centering
\includegraphics[scale=0.6]{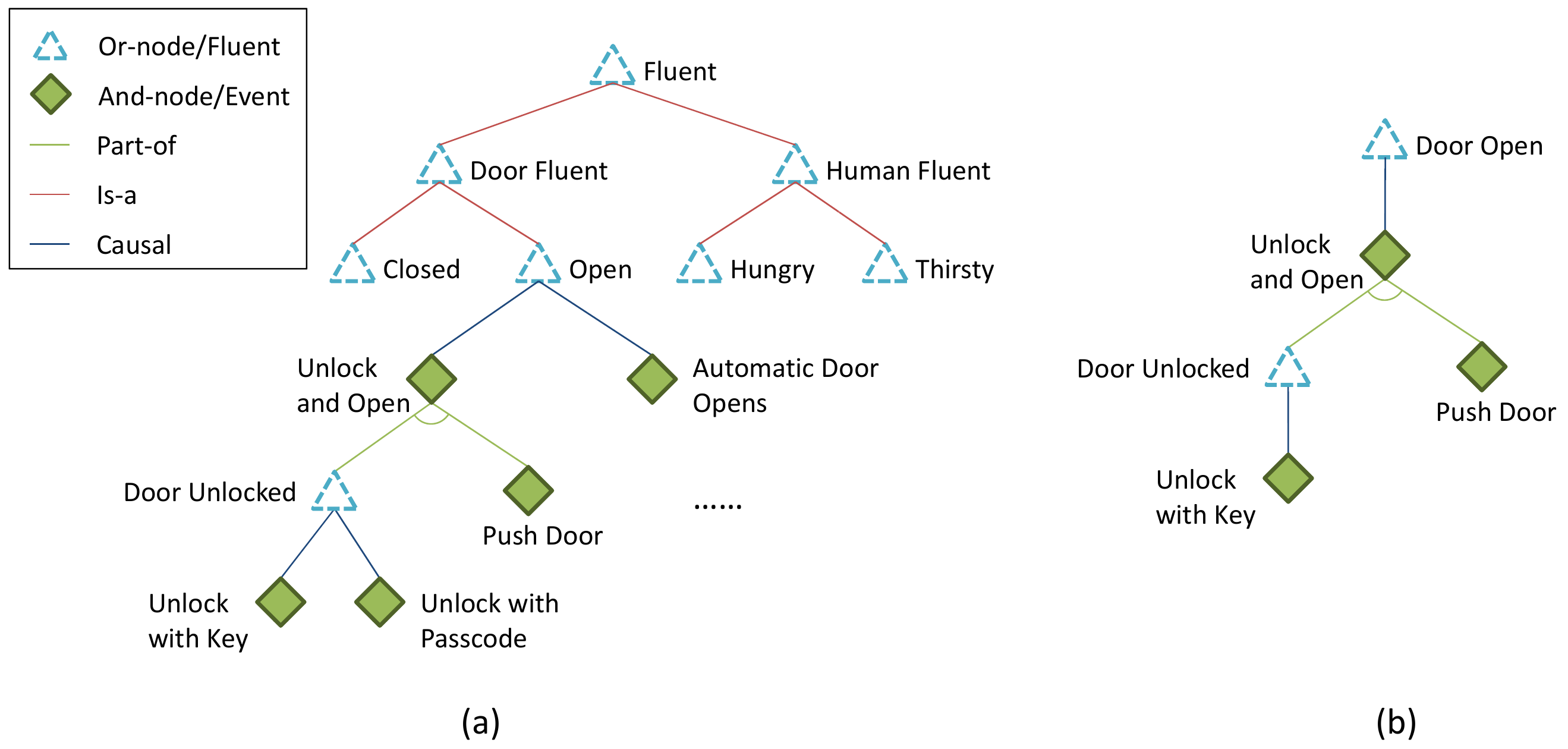}
\caption{(a) An example causal And-Or graph (C-AOG). An And-node represents the composition of its child nodes (conditions and sub-events); an Or-node represents a selection among its child nodes; a leaf node represents an event or a fluent.
(b) An example causal parse graph, which is a realization of the C-AOG by making selections at Or-nodes.}
\label{fig:c-aog}
\end{figure*}

A causal And-Or graph (C-AOG) \cite{Fire13uc} captures the knowledge of the \texttt{Causal} relation between events and fluent changes. Figure \ref{fig:c-aog}(a) shows an example C-AOG. 
An And-node in a C-AOG represents a composition of conditions and events that can cause a fluent change. For example, to cause a door to open, it can be the case that the door is unlocked and someone pushes the door.
An Or-node in a C-AOG represents alternative causes that can result in a fluent change. For example, the cause that a door is opened can be either that ``someone unlocks and pushes the door'' or that ``the door is automatic and someone walks close''.
A leaf node in a C-AOG is either an event or a fluent.

\subsubsection{Joint S/T/C-AOG}\label{sec:rep:aog:j}
\begin{figure}
\centering
\includegraphics[scale=0.55]{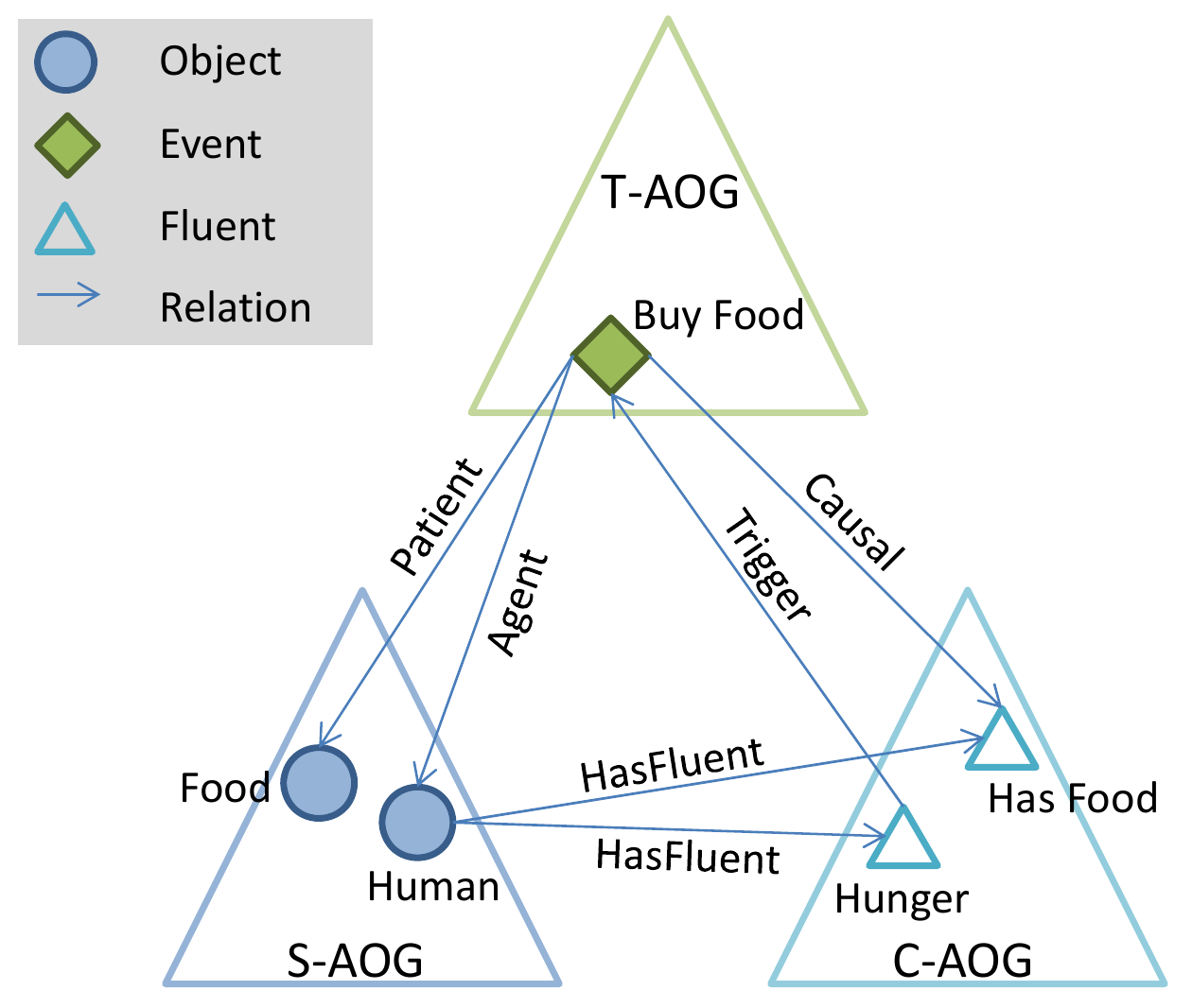}
\caption{An example of the relations between S-AOG, T-AOG and C-AOG.}
\label{fig:j-aog}
\end{figure}

Although S-AOG, T-AOG and C-AOG have been proposed and investigated separately before, in this work we show for the first time that the three types of AOGs are interrelated and form a joint S/T/C-AOG, as shown in Figure \ref{fig:j-aog}. An action represented by a leaf node in the T-AOG is connected via the \texttt{Agent}, \texttt{Patient} and \texttt{Location} relations to the action initiator, target and location, which are objects modeled in the S-AOG. An event that has a \texttt{Causal} relation with a fluent change in the C-AOG is itself modeled by the T-AOG. There may also be a \texttt{Trigger} relation between a fluent in the C-AOG and an event in the T-AOG. Finally, there are \texttt{HasFluent} relations between objects in the S-AOG and fluents in the C-AOG.
The connections across the three types of AOGs enable the propagation of information between them during parsing, which improves parsing accuracy. Connecting the three types of AOGs also leads to a more coherent and comprehensive joint interpretation, which facilitates subsequent applications such as query answering of who, what, when, where and why.

\subsection{Parse graphs}\label{sec:rep:pg}
We represent an interpretation of the input video and/or text with a graphical representation called a parse graph. 
A parse graph is a realization of the S/T/C-AOG by selecting at each Or-node one of the alternative configurations. For example, Figure \ref{fig:s-aog}(b), \ref{fig:t-aog}(b) and \ref{fig:c-aog}(b) show respectively a valid spatial, temporal and causal parse graph as realizations of the S/T/C-AOGs given in Figure \ref{fig:s-aog}(a), \ref{fig:t-aog}(a) and \ref{fig:c-aog}(a).
A parse graph can be represented as a labeled directed graph.
Each node in a parse graph represents an entity whose label indicates the semantic type of the entity. A node may also have a set of spatial-temporal annotations indicating the spatial-temporal coordinate or range of the entity represented by the node. A parse graph may also contain a special type of nodes representing attribute values in the form of strings or numbers. 
A directed edge between two nodes in a parse graph represents a relation between the two entities denoted by the two nodes. The label on the edge indicates the type of the relation.
We adopt the \emph{open world assumption} \cite{Russel03aima} which states that a parse graph may not be a complete interpretation of the scene and any statement not included or implied by the parse graph can be either true or false.

Figure \ref{fig:suv_ex}(b) and \ref{fig:foodtruck_ex}(b) show the parse graphs that represent the interpretations of the videos and texts shown in Figure \ref{fig:suv_ex}(a) and \ref{fig:foodtruck_ex}(a).
Note that for better visualization of the parse graph, we use different node shapes to denote different types of entities, and use different edge colors to denote different types of relations; for nodes with temporal annotations (e.g., events), we lay them out in accordance with the time line; for a fluent that may have its value changed over time, 
we use levels to indicate the value of the fluent at each time point.

We infer parse graphs from video and text through maximizing the posterior probability that we shall discuss in the following few sections. As an explicit semantic representation of the video and text, our parse graphs facilitate subsequent applications such as text generation and query answering.
In addition, our parse graph can be naturally represented as a Resource Description Framework (RDF) data model \cite{RDF04}, one of the standard knowledge representations in the semantic web. This allows easy distribution of the parse graph data in the semantic web and enables the use of a multitude of approaches and tools for the editing, storage, inference and retrieval of the data.

\subsection{Probabilistic Modeling}\label{sec:rep:prob}
In this section we define a probabilistic distribution over parse graphs to account for their stochasticity. Since the valid parse graphs are specified by an AOG, we assign energy terms to the AOG to make it a probabilistic model of parse graphs (so an AOG embodies a stochastic grammar). First, at each Or-node of the AOG we specify 
an energy that indicates how likely each alternative configuration under the Or-node is selected in a parse graph. Second, at each And-node of the AOG we specify an energy for each relation between the child nodes of the And-node that captures the uncertainty of the relation. These energy terms can be either manually specified by domain experts or learned from data \cite{Si2012la,Si11ule,Fire13uc}.
The energy of a parse graph is thus determined by the energy of configuration selections made at the Or-nodes and the energy of the relations between the child nodes of the And-nodes contained in the parse graph. Let $V^{or}(pg)$ be the set of Or-nodes in $pg$, $E_{or}(v)$ the energy associated to the configuration selected at the Or-node $v$, $R(pg)$ the set of relations specified at the And-nodes in $pg$, and $E_R(r)$ the energy associated with the relation $r$. The energy of a parse graph is defined as:
\begin{equation}\label{eq:E_1aog}
	E(pg) = \sum_{v \in V^{or}(pg)} E_{or}(v) + \sum_{r \in R(pg)} E_R(r)
\end{equation}
We also allow part of a parse graph to be missing with a penalty to accommodate incomplete observation or description.

As introduced in section \ref{sec:rep:aog}, we use a joint S/T/C-AOG to define the valid parse graphs. The energy of a parse graph in our framework is therefore the summation of four energy terms.
\begin{equation}\label{eq:E_aog}
	E_{\mathrm{STC}}(pg) = E_\mathrm{S}(pg) + E_\mathrm{T}(pg) + E_\mathrm{C}(pg) + \sum_{r \in R^*(pg)} E_R(r)
\end{equation}
where $E_\mathrm{S}(pg)$, $E_\mathrm{T}(pg)$ and $E_\mathrm{C}(pg)$ are the energy terms defined by the S-AOG, T-AOG and C-AOG respectively according to Eq.\ref{eq:E_1aog}, $R^*(pg)$ is the set of relations that connect the three types of AOGs (as discussed in section \ref{sec:rep:aog:j}) and are not contained in any of the AOGs, and $E_R(r)$ is the energy associated with the relation $r$.

The prior probability of a parse graph $pg$ is then defined as:
\begin{equation}\label{eq:prior}
	P(pg) = \frac{1}{Z} e^{-E_{\mathrm{STC}}(pg)}
\end{equation}
where $Z$ is the normalization factor.
Eq.\ref{eq:prior} defines one of the four factors in the posterior probability of the parse graphs given the input video and text (Eq.\ref{eq:obj}). In the next three sections, we shall introduce and discuss the other three factors which are involved respectively in video parsing, text parsing and joint inference.

\section{Video Parsing}\label{sec:video}
In video parsing, we aim to detect objects, scenes and events that form an interpretation of the input video. Since our prior knowledge of objects, scenes and events is encoded in the S/T/C-AOG, we regard the S/T/C-AOG as the prior of the video parse graph $pg_{vid}$ and want to optimize the posterior of $pg_{vid}$. Our objective energy function for video parsing is:
\begin{equation}\label{eq:E_v}
	E_v(pg_{vid}) = E_{\mathrm{STC}}(pg_{vid}) - \log p(vid|pg_{vid})
\end{equation}
where $E_{\mathrm{STC}}$ is defined in Eq.\ref{eq:E_aog}, and $p(vid|pg_{vid})$ is the probability of the input video $vid$ given the objects, scenes and events specified in the parse graph $pg_{vid}$.
Since video parsing often contains a significant level of uncertainty, we output a set of candidate video parse graphs with low energy instead of a single best parse graph. 
To justify the use of the objective energy function $E_v$ as defined in Eq.\ref{eq:E_v}, note that if we assume that the set of nodes and edges in the joint parse graph $pg_{jnt}$ is a superset of that in the video parse graph $pg_{vid}$, then it can be shown that $\exp(-E_v(pg_{vid}))$ is a factor of the joint posterior probability $P(pg_{jnt}, pg_{vid}, pg_{txt}|vid,txt)$ (Eq.\ref{eq:obj}); therefore, a video parse graph that has a high energy according to Eq.\ref{eq:E_v} would very likely lead to a low joint posterior probability, so we use Eq.\ref{eq:E_v} to prune such video parse graphs.

\subsection{Spatial parsing}
We first perform spatial parsing on each video frame following the approach proposed in \cite{Zhao11ip} called hierarchical cluster sampling. The approach consists of two stages. In the first stage, the approach performs bottom-up clustering in which lower-level visual entities (e.g., line segments) of a video frame are composed into possible higher level objects according to the S-AOG. To keep the number of candidate objects tractable, compositions with high energy are pruned. In the second stage, the parser applies the Metropolis-Hastings algorithm in the space of spatial parse graphs. The sampler performs two types of operations to change the current spatial parse graph: the first is to add a candidate object composed in the first stage into the parse graph, where the proposal probability is defined based on the energy of the candidate object as well as the compatibility of the object with the rest of the parse graph; the second type of operation is to randomly prune the current parse graph. At convergence the approach outputs the optimal spatial parse graph.

\subsection{Temporal parsing}
We perform temporal parsing following the approach proposed in \cite{Pei11pv}, which is based on the Earley parser \cite{Earley83ae}.
The input video is divided into a sequence of frames. The agents, objects and fluents in each frame are identified using the spatial parser and special detectors. The temporal parser reads in frames sequentially, and at each frame maintains a so-called state set that contains pending derivations of And-nodes that are consistent with the input video up to the current frame (i.e., for each And-node in the state set, only a subset of its child nodes have been detected from the video up to the current frame).
At the beginning of parsing, the And-nodes at the top level of the T-AOG are added into the state set of frame 0, with all their child nodes pending.
With each new frame read in, three basic operations are iteratively executed on the state set of the new frame. The \emph{prediction} operation adds into the state set new And-nodes that are expected according to the existing pending derivations in the state set. The \emph{Scanning} operation checks the detected agents, objects and fluents in the new frame and advances the pending derivations in the state set accordingly. The \emph{Completion} operation identifies And-nodes whose derivations are complete and advances the pending derivations of their parent And-nodes. During this process, we prune derivations that have high energies to make the sizes of state sets tractable.
After all the frames are processed, a set of candidate parses of the whole video can be constructed from the state sets.

\subsection{Causal parsing}
After all the events are detected in temporal parsing, we then perform causal parsing. For each fluent change detected in the video using special detectors, we collect the set of events that occur within a temporal window right before the fluent change and run the Earley parser again based on the sub-graph of the C-AOG rooted at the detected fluent change.

\begin{figure}\centering
  \includegraphics[scale=0.6]{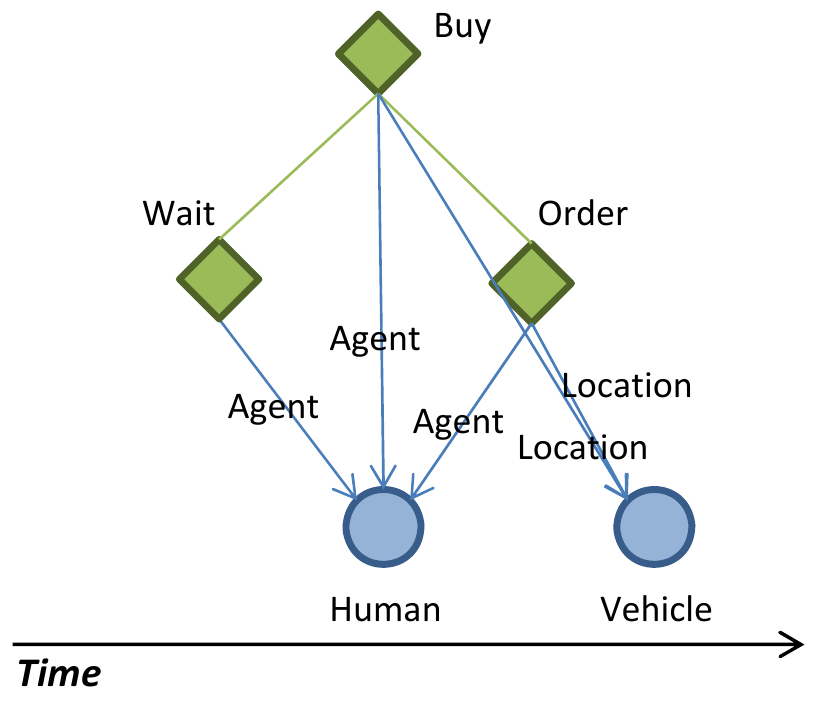}\\
  \caption{A candidate parse graph of the video in our example input shown in Figure \ref{fig:foodtruck_ex}(a).}\label{fig:pgv}
\end{figure}

After spatial, temporal and causal parsing is done, we combine the resulting spatial, temporal and causal parse graphs into a complete video parse graph. Figure \ref{fig:pgv} shows a parse graph of the example input video shown in Figure \ref{fig:foodtruck_ex}(a).

\section{Text Parsing}\label{sec:text}
In parallel to video parsing, we perform text semantic parsing to convert the input text description into the text parse graph $pg_{txt}$. Our semantic parsing approach is similar to many existing information extraction approaches, but our objective is distinctive in two aspects. First, we focus on the domain of narrative event descriptions to extract information about objects, events, fluent-changes and the relations between these entities. Important relations in this domain include spatial, temporal, functional and causal relations. Second, information extracted from the text is expressed using the same set of semantic types and parse graph representation that are also used in video parsing. This allows us to connect the two parse graphs and perform joint inference.
We assume that text descriptions are unambiguous and therefore only a single text parse graph is produced from the input text. However, our approach can also handle text ambiguity and produce multiple candidate text parse graphs with different probabilities.
Our semantic parsing of text is a processing pipeline consisting of four main steps: text filtering, part-of-speech tagging and dependency inference, dependencies filtering, and parse graph generation, which will be introduced in the following four subsections. In the case where the text descriptions are accompanied with time stamps or durations, we add temporal annotations to the corresponding event nodes in the resulting parse graph.
We will use the parsing process of the sentence ``Tom is buying food'' as a running example in the rest of this section.

\subsection{Text Filtering}
This is a preprocessing step that locates and labels certain categories of words or phrases in the input text to facilitate subsequent text parsing steps. It first performs the named entity recognition (NER) to identify text elements related to names of persons, time, organizations and locations. Existing NER tools such as Stanford NER \cite{Finkel05in} can be used. Second, it recognizes certain compound noun phrases that refer to a single entity (e.g., ``food truck'' and ``vending machine''). The words in a compound noun phrase are then grouped together into a single word. Currently this is done by utilizing the chunker tool in Apache OpenNLP \cite{OpenNLP}.
Our example sentence is not changed in this step.

\subsection{Part-of-Speech Tagging and Dependency Inference}
This step performs part-of-speech (POS) tagging and syntactic dependency parsing using the Stanford Lexicalized Parser \cite{SP}. During POS tagging, each word is assigned a grammatical category such as noun, verb, adjectives, adverb, article, conjunct and pronoun. The tagger uses the Penn treebank tag set which has 45 tags. The parser then performs dependency analysis to identify relations between the words in the sentence. It generates a dependency graph in which each node is a word and each directed edge represents a basic grammatical relation between the words. The Stanford Dependencies \cite{SPM} include more than 50 types of grammatical relations such as subject, modifier, complement, direct and direct object. For example, the following dependency identifies Tom as the subject of the buying event. 
\begin{center}
    nsubj(buying,Tom).
\end{center}
Together with POS labels, these dependency relations allow us to extract information from the text more easily.

\begin{figure}\centering
  \includegraphics[scale=0.6]{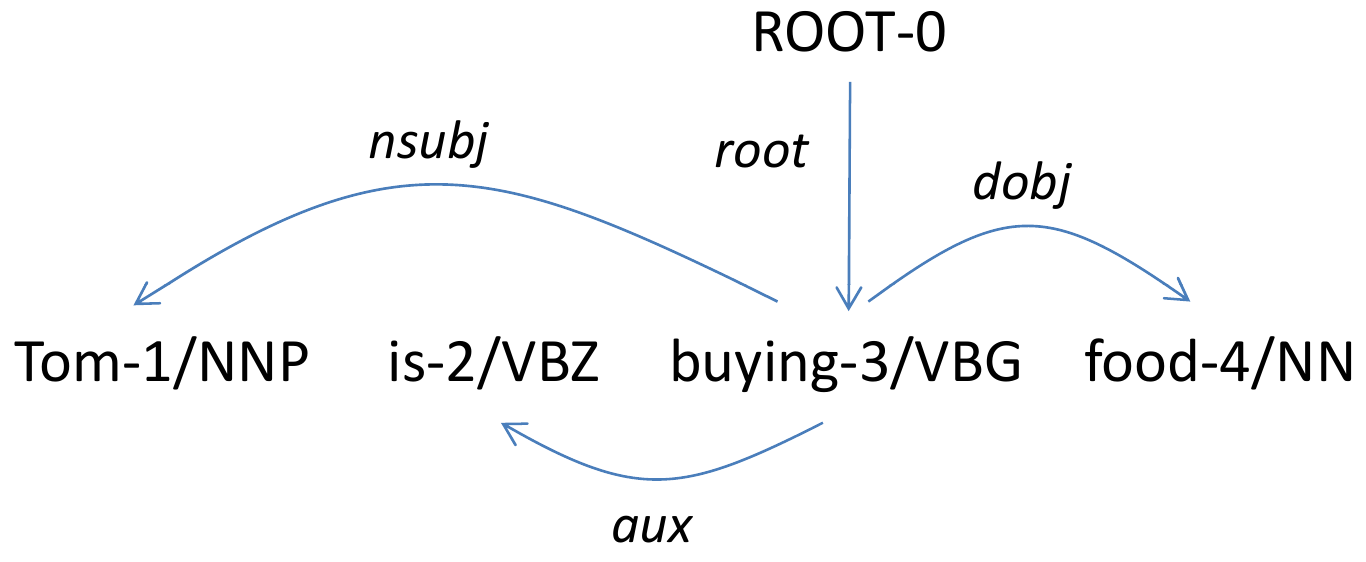}\\
  \caption{An example dependency tree from step 2 of text parsing.}\label{fig:pgt_s2}
\end{figure}

For our example sentence, the result of this step is shown in Figure \ref{fig:pgt_s2}.

\subsection{Dependencies Filtering}
In this step, we map each word in the sentence to an entity type defined in the ontology or to a literal type.
Examples of such mappings are shown in Table \ref{tab:sem_map}. This step is currently implemented using a look-up table but can also be automated by utilizing the WordNet dictionary \cite{WordNet}.

\begin{table}
  \centering
  \begin{tabular}{|p{20ex}|p{38ex}|}
    \hline
    \textbf{Entity type or literal type} & \textbf{Words in the input text} \\\hline
    \hline
    Human & person \textbar\ individual \textbar\ somebody \textbar\ \ldots \\\hline
    Beverage & drink \textbar\ beverage \textbar\ \ldots \\\hline
    Buying & buying \textbar\ purchasing \textbar\ \ldots \\\hline
    Vending\_machine & vending\_machine \textbar\ slot\_machine \textbar\ coin\_machine \textbar\ \ldots \\\hline
    Number & one \textbar\ two \textbar\ three \textbar\ \ldots \\\hline
    Color & red \textbar\ blue \textbar\ \ldots \\\hline
    Aux & is \textbar\ are \\\hline
  \end{tabular}
  \smallskip
  \caption{Example mappings between words and entity types or literal types.}\label{tab:sem_map}
\end{table}

In some cases, it is important to check the POS tag of a word to determine the correct entity type. For example, in the following two sentences:
\begin{center}
    The door is open(/JJ).\\
    They open(/VBP) the door.
\end{center}
the word ``open'' in the first sentence refers to a fluent state, while in the second sentence it refers to an action. The POS labels ``JJ'' and ``VBP'' allow us to map the words to the correct entity types. 
Note that since we focus on a small set of entity types as specified in our ontology, words that can denote multiple entity types under the same POS tags are rare.

For our example sentence, the resulting dependency tree of this step is:
\begin{center}\renewcommand{\arraystretch}{1}
\begin{tabular}{l}
    nsubj(Buying, Human\_name)\\
    aux(Buying, Aux)\\
    root(ROOT-0, Buying)\\
    dobj(Buying, Food)
\end{tabular}
\end{center}
We further replace each entity type with its top level supertype specified in the ontology. In our example, ``Buying'' is replaced with ``Event'' and ``Food'' is replaced with ``Object''.

Compared with the original dependency tree which describes the relations between words, we have now added the entity types of these words. This provides additional cues to infer more precise semantic relations between the entities denoted by the words. For instance, from the above example, the dependency ``nsubj(Event, Human\_name)'' implies that the person (Tom) is the agent of the event (buying). Without the entity types, the dependency relation ``nsubj'' could be ambiguous, e.g., ``nsubj(teacher, Tom)'' from the sentence ``Tom is a teacher'' refers to an occupation attribute and not an agent-event relation.

\subsection{Parse Graph Generation}
\begin{figure*}\centering
  \includegraphics[scale=0.6]{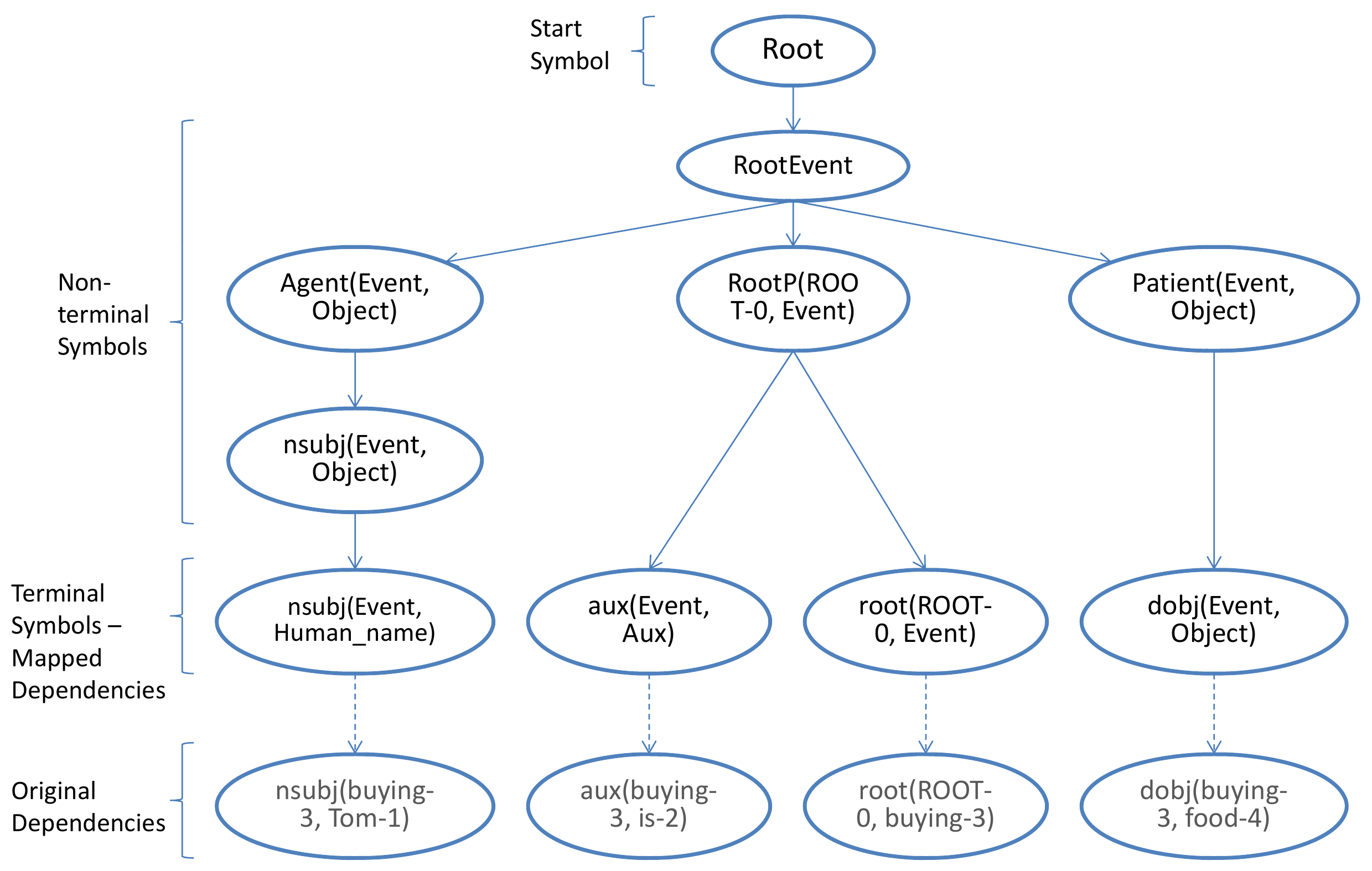}\\
  \caption{An illustration of step 3 and 4 of text parsing: dependencies filtering of step 3 is shown in the bottom half of the figure; attribute grammar parsing of step 4 is shown in the top half of the figure.}\label{fig:pgt_s4}
\end{figure*}

In order to extract semantic relations from dependencies more accurately, we need to consider not only individual dependencies but also their contexts (the semantic types of the words and the neighboring dependencies). Therefore, we design an attribute grammar to parse the set of dependencies and infer the semantic relations. The final text parse graph is then generated from the semantic relations inferred in this step as well as the entity types produced from the previous step.

In the attribute grammar, the terminal symbols are the mapped dependencies from the previous step. A set of production rules describe how the terminal symbols can be generated from ROOT, the start symbol.
The Earley parsing algorithm \cite{Earley83ae,Hakeem09sv} is used to parse the dependencies using the attribute grammar, which employs top-down dynamic programming to search for a valid set of production rules that generate the dependencies. 
The parse tree from the attribute grammar for our example sentence is shown in Figure \ref{fig:pgt_s4}.

\begin{figure}\centering
  \includegraphics[scale=0.6]{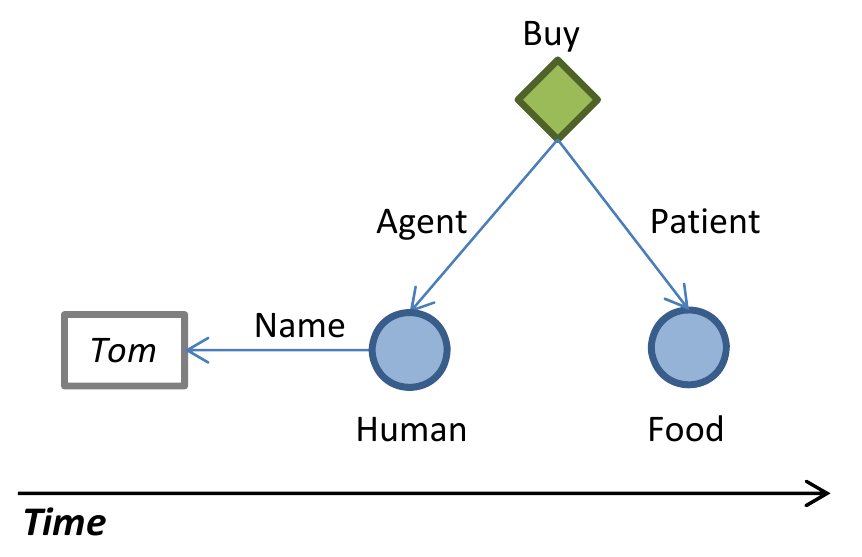}\\
  \caption{The parse graph of the example input text ``Tom is buying food''.}\label{fig:pgt}
\end{figure}

In each node of the parse tree, information from the input sentence is organized into an attribute-value structure, which may include attributes such as the event or action type, person's name, etc. 
Such information structures are propagated from the lower-level nodes to the higher-level nodes of the parse tree via the attribute functions associated with the production rules of the attribute grammar. 
For example, in the following production rule (which is used in the left-most branch of the parse tree in Figure \ref{fig:pgt_s4}):
\begin{quote}
nsubj(Event$_1$, Object)\\
\phantom{xxxx}$\rightarrow$ nsubj(Event$_2$, Human\_name)
\end{quote}
the associated attribute functions are:
\begin{quote}
Event$_1$.$*$ := Event$_2$.$*$ \\
Object.$*$ := Human\_name.$*$ \\
Object.type := Human
\end{quote}
where ``$*$'' is a wild-card character that stands for any subtext of the attribute name. 
After all the information from the input sentence is propagated to the top-level nodes in the parse tree of the attribute grammar, the final text parse graph is then generated based on these top-level nodes. For example, the top-level node \texttt{Agent(Event,Object)} in Figure \ref{fig:pgt_s4} contains the following information:
\begin{quote}
Event.id = ``buying-3'' \\
Event.type = Buy \\
Object.id = ``Tom-1'' \\
Object.type = Human \\
Object.name = ``Tom''
\end{quote}
from which we can construct a part of the text parse graph containing the \texttt{Agent} relation between the buying event and the human as well as the name attribute of the human.
The final text parse graph of the example sentence is shown in Figure \ref{fig:pgt}.

In addition to extracting the \texttt{Agent} and \texttt{Patient} relations as shown in the running example, by using the attribute grammar we also extract the spatial, temporal and causal relations as well as functional properties of objects based on the function words in the input sentence. Examples of such relations and the function words used to extract such relations are shown in Table \ref{tab:function_word}. Some function words, such as ``at'', ``from'' and ``to'', have multiple meanings, but their semantic context in the input sentence allows the correct relations to be inferred during parsing. 

\begin{table}
  \centering
  \begin{tabular}{|l|p{25ex}|p{22ex}|}
    \hline
    \textbf{Category} & \textbf{Relations} & \textbf{Function words} \\\hline
    \hline
    Spatial & location, locationFrom, locationTo & at, on, in, from, to, into \\\hline
	Temporal & timeAt, timeFrom, timeTo, timeBefore, timeAfter & at, from, to, before, after \\\hline
	Causal & causal, trigger & therefore, because \\\hline
	Functional & occupation, religion, race & is a \\\hline
  \end{tabular}
  \smallskip
  \caption{Examples of relations extracted from text and the corresponding function words.}\label{tab:function_word}
\end{table}

\section{Joint Inference}\label{sec:joint}
In joint inference, we construct the joint parse graph from the video and text parse graphs produced by the first two modules.
Although spatial-temporal-causal parsing of videos and semantic parsing of text have been studied in the literature, this is the first time to integrate them together.

\subsection{Challenges in Joint Inference}
The joint parse graph shall not be a simple union of the video parse graph and the text parse graph. We need to consider the following three difficulties.
\begin{enumerate}
    \item \textbf{Coreference}. The entities and relations covered by the two input parse graphs may be overlapping. Figure \ref{fig:match} shows such an example. We need to identify and merge the overlapping entities and relations. One challenge is that the same entity or relation might be annotated with different semantic types in the two parse graphs. For example, in Figure \ref{fig:match} the food truck entity is annotated with ``Vehicle'' in the video parse graph and ``Food\_Truck'' in the text parse graph. In addition, the spatial-temporal annotations of the same entity may not be exactly the same in the two parse graphs. As another challenge, for each entity there might be multiple candidate matching targets. For example, a person mentioned in the text may be matched to a few different persons in the video.
	\begin{figure}\centering
	  \includegraphics[scale=0.6]{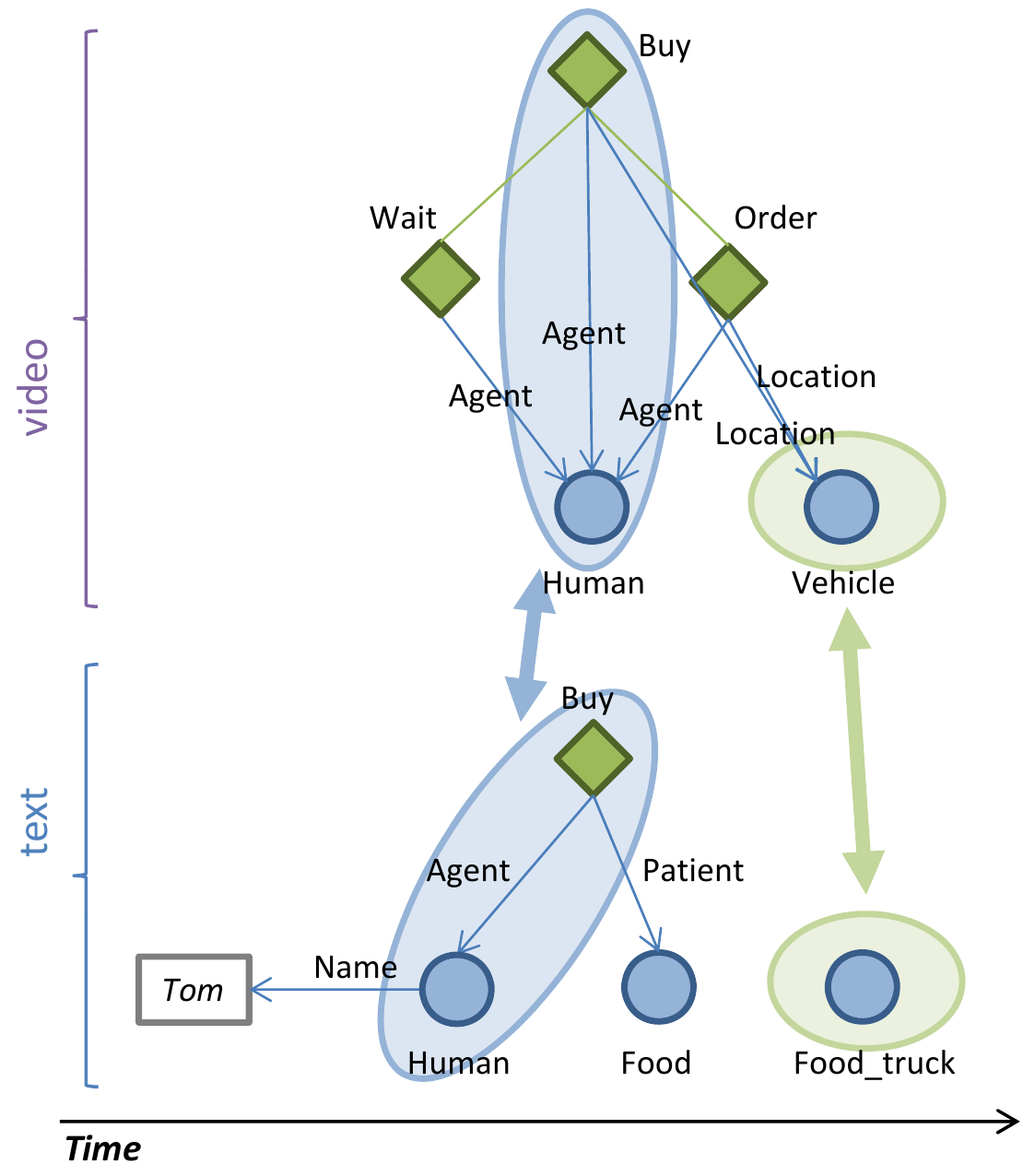}\\
	  \caption{An example of overlapping video and text parse graphs. The ovals and bidirectional arrows mark the subgraphs that represent the same entities and relations.}\label{fig:match}
	\end{figure}

    \item \textbf{Missing information}. Some information is not explicitly presented in the input video and text, but may be deduced using the prior knowledge encoded in the S/T/C-AOG. Such information is necessary to fill in the gap between the input video and text or is useful in downstream tasks like query answering. So we should include such information in the joint parse graph. Figure \ref{fig:add} shows an example in which the fluent ``Has\_Food'' is neither observable in the video nor stated in the text, but can be deduced from the event of buying food.
	\begin{figure}\centering
	  \includegraphics[scale=0.6]{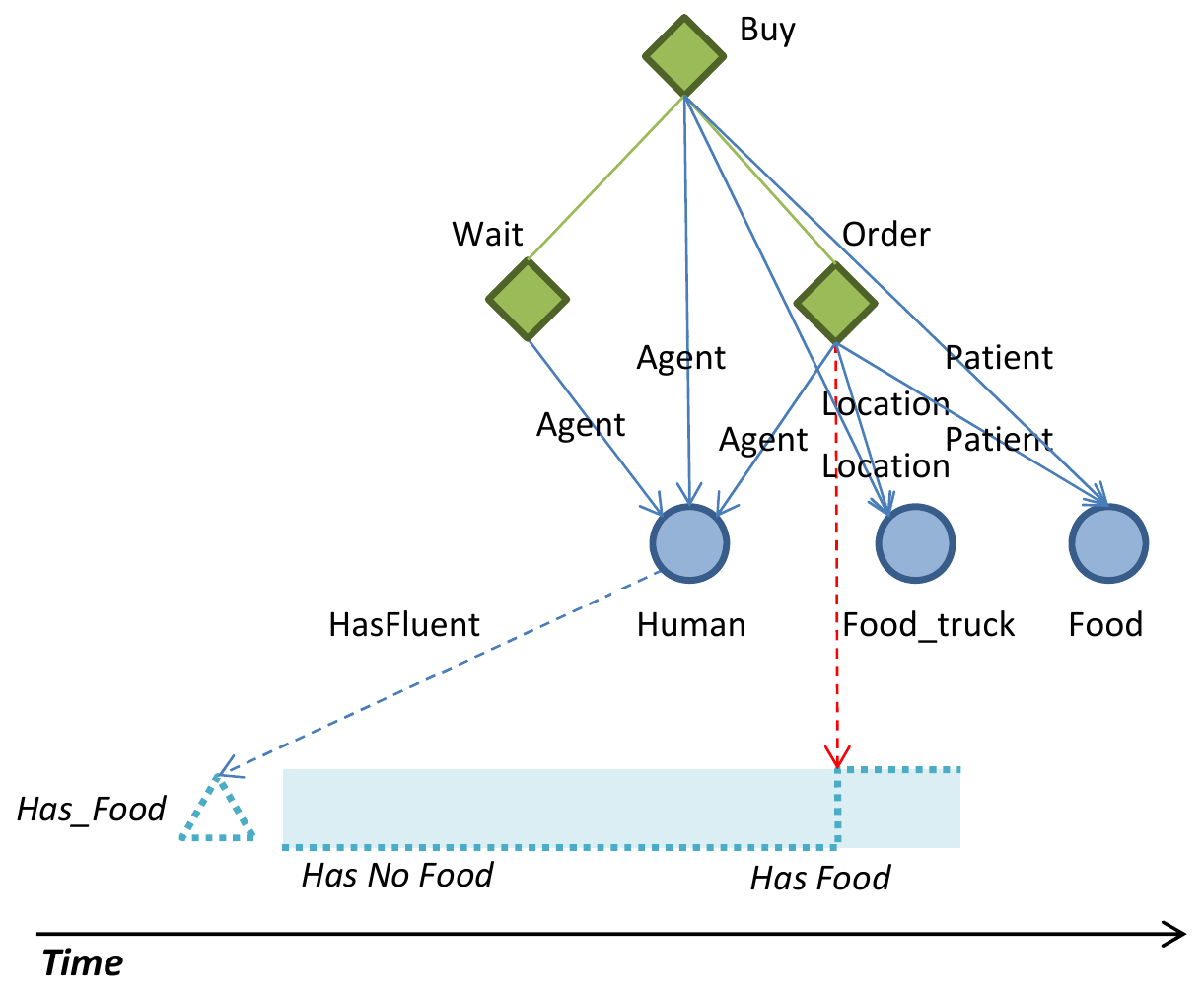}\\
	  \caption{An example of information that can be deduced from the parse graph. The dashed shapes and arrows represent inferred entities and relations.}\label{fig:add}
	\end{figure}

    \item \textbf{Conflicts}. The parse graphs from the input video and text may contain conflicting information. Such conflicts manifest themselves once overlapping entities and relations from the two parse graphs are merged. Figure \ref{fig:conflict} shows an example parse graph that contains conflicting information. We need to detect such conflicts using the prior knowledge and correct them in the joint parse graph.
	\begin{figure}\centering
	  \includegraphics[scale=0.6]{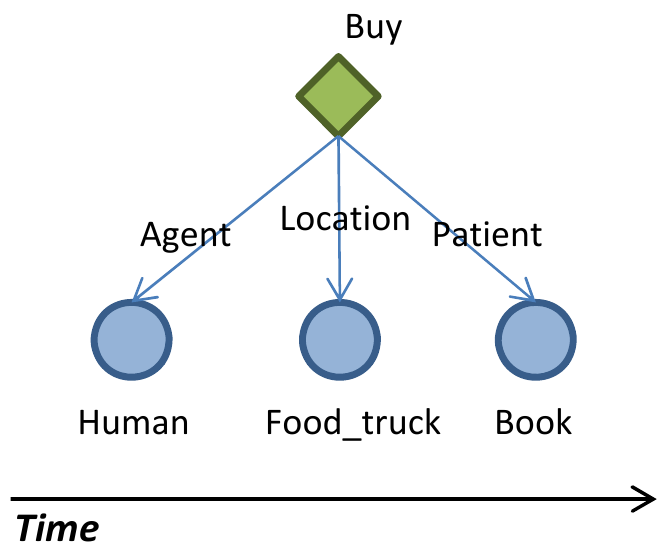}\\
	  \caption{An example of conflicting information in the parse graph. Here the conflict is between ``buy book'' and ``buy from food truck''.}\label{fig:conflict}
	\end{figure}
\end{enumerate}

\begin{figure}\centering
  \includegraphics[scale=0.5]{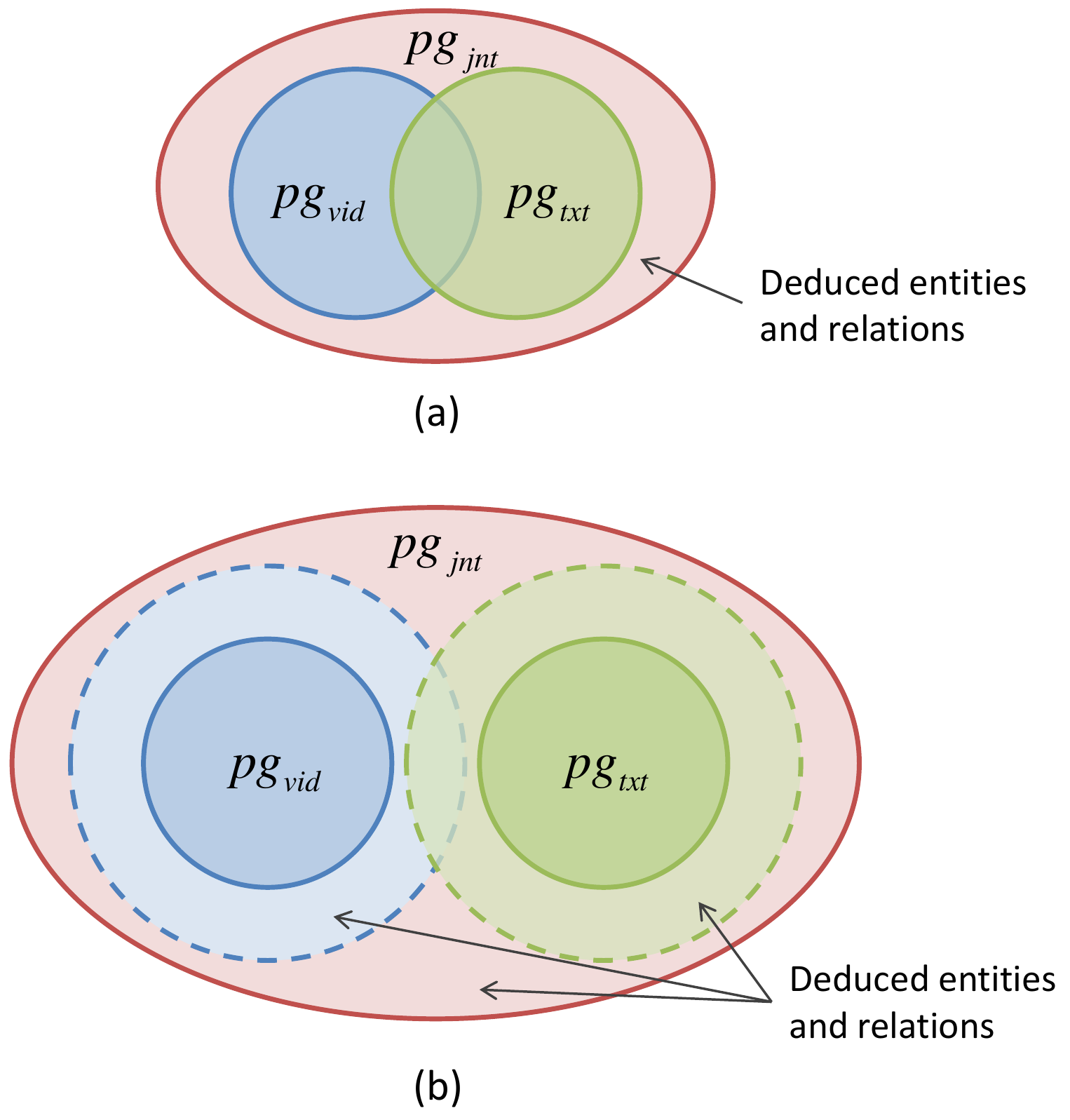}\\
  \caption{A schematic diagram of how the joint parse graph $pg_{jnt}$ is constructed from the video parse graph $pg_{vid}$ and text parse graph $pg_{txt}$, when (a) there is overlapping between the video and text parse graphs, and (b) the video and text parse graphs have no overlapping but become overlapped after some steps of deduction.}\label{fig:joint_inf_diagram}
\end{figure}

Figure \ref{fig:joint_inf_diagram} shows a schematic diagram of how the joint parse graph is constructed from the video and text parse graphs, which demonstrates the three difficulties in joint inference. 
There are two scenarios. In the first scenario (Figure \ref{fig:joint_inf_diagram}(a)) where there is significant overlapping between the video and text parse graphs, we first need to identify the overlapping and solve the coreference problem, and then we can deduce missing information and correct conflicts to obtain a more accurate and comprehensive joint parse graph. In the second scenario (Figure \ref{fig:joint_inf_diagram}(b)) where there is no overlapping between the video and text parse graphs, at first no coreference can be identified to connect the two parse graphs; only after the two parse graphs are expanded by deducing implicit information do they become overlapped, and then we can connect them by solving coreference and produce a coherent joint parse graph.

To solve the three difficulties, we propose three types of operators: \emph{matching}, which matches and merges subgraphs of the video and text parse graphs to solve the coreference problem; \emph{deduction}, which inserts implicit information to fill in the potential gap between video and text and makes the joint parse graph more comprehensive; and \emph{revision}, which resolves possible conflicts between video and text.
In the rest of this section, we first define the conditional probability $P(pg_{vid}, pg_{txt} | pg_{jnt})$ in section \ref{sec:joint:lh}, which specifies the relation between the joint parse graph and the video and text parse graphs; then we introduce the joint inference algorithm in section \ref{sec:joint:alg}, which takes the union of the video and text parse graphs as the initial joint parse graph and iteratively changes it using the three types of operators to maximize the joint posterior probability.

\subsection{Relating the three parse graphs}\label{sec:joint:lh}
To relate the video parse graph $pg_{vid}$, the text parse graph $pg_{txt}$ and the joint parse graph $pg_{jnt}$, we define the conditional probability $P(pg_{vid},pg_{txt} | pg_{jnt})$. By multiplying it with the prior probability of the joint parse graph $P(pg_{jnt})$ (based on Eq.\ref{eq:prior}), we get the joint probability of the three parse graphs, which is used to guide our joint inference.
Intuitively, the conditional probability $P(pg_{vid},pg_{txt} | pg_{jnt})$ models the generation of the video and text parse graphs given the joint parse graph, by considering how likely different types of entities and relations in the joint parse graph would be included in the video and text parse graphs, while penalizing the changes of the semantic, temporal and spatial annotations of these entities and relations in the video and text parse graphs. 
We do not assume independence of the generation of the video parse graph and that of the text parse graph, because in many scenarios they are indeed generated jointly, e.g., in TV news reporting the news editor would choose the footage and narrative that cover the same aspect of the whole news story.
To make the maximum a posteriori estimation in joint inference tractable, we make the assumption that the inclusion and change of each entity or relation is independent of that of any other entity or relation in the parse graph, so we can factorize the conditional probability as defined below.
\begin{eqnarray}\label{eq:lh}
    P(pg_{vid},pg_{txt} | pg_{jnt}) &=& \prod_{x \in pg_{jnt}} \frac{1}{Z_x} e^{-E(x)} \\
    && \times \frac{1}{Z_\psi} e^{-\alpha_\psi \|(pg_{vid} \cup pg_{txt}) \setminus pg_{jnt}\|} \nonumber
\end{eqnarray}
where $x$ ranges over all the entities and relations in the joint parse graph $pg_{jnt}$, $Z_x$ and $Z_\psi$ are normalization factors, $\alpha_\psi$ is a constant representing the penalty of having an entity or relation in the video or text parse graph that is not contained in the joint parse graph, and $E(x)$ is defined as follows.
\begin{eqnarray*}
	\lefteqn{E(x) =} \\
	&& \left\{ \begin{array}{ll}
		\alpha_v(x) + d(x_j,x_v) & \textrm{if $x \in pg_{vid} \setminus pg_{txt}$} \\
		\alpha_t(x) + d(x_j,x_t) & \textrm{if $x \in pg_{txt} \setminus pg_{vid}$} \\
		\alpha_{vt}(x) + d(x_j,x_v) + d(x_j,x_t) & \textrm{if $x \in pg_{vid} \cap pg_{txt}$} \\
		\alpha_\phi(x) & \textrm{if $x \notin pg_{vid} \cup pg_{txt}$} \\
	\end{array}\right.
\end{eqnarray*}
We enforce the constraint that a relation in the joint parse graph can be contained in the video or text parse graph only if the two entities connected by the relation are also contained in the video or text parse graph.
The notations used in the energy function are explained below.
\begin{itemize}
\item $\alpha_v(x)$ is the energy that $x$ is contained in the video parse graph $pg_{vid}$ but not in the text parse graph $pg_{txt}$. Ideally, this energy shall be set to a low value for elements that represent visible low-level details of objects, scenes and events, for example, the atomic actions of a drinking event (e.g., reaching to, lifting, holding and putting down a cup of water), which are seldom mentioned in the text description.
\item $\alpha_t(x)$ is the energy that $x$ is contained in the text parse graph $pg_{txt}$ but not in the video parse graph $pg_{vid}$. Ideally, this energy shall be set to a low value for elements representing objects and events or their attributes that are either not visible in the video (e.g., a person's name, whether he is hungry) or hard to detect by the video parser (e.g., a person's gender can be hard to identify in a low resolution surveillance video).
\item $\alpha_{vt}(x)$ is the energy that $x$ is contained in both the video parse graph $pg_{vid}$ and the text parse graph $pg_{txt}$. Ideally, this energy shall be set to a low value for elements representing visible high-level objects, scenes and events.
\item $\alpha_\phi(x)$ is the energy that $x$ is contained in neither the video parse graph $pg_{vid}$ nor the text parse graph $pg_{txt}$. Ideally, this energy shall be set to a low value for elements representing low-level details of objects, scenes and events that are invisible or hard to detect in the video.
\item $x_j$, $x_v$ and $x_t$ are the corresponding entities or relations of $x$ in the joint parse graph $pg_{jnt}$, video parse graph $pg_{vid}$ and text parse graph $pg_{txt}$ respectively. 
\item $d$ is a distance measure that combines the semantic, temporal and spatial distances as defined below. It models the difference in the annotations of the entity or relation $x$ in the joint parse graph and the video or text parse graph.
\begin{equation}\label{eq:dist}
    d(x, y) = \alpha_o d_o(x,y) + \alpha_t d_t(x,y) + \alpha_s d_s(x,y)
\end{equation}
where $\alpha_o, \alpha_t$ and $\alpha_s$ are constant weights, and the three distance measures $d_o(x,y), d_t(x,y)$ and $d_s(x,y)$ are explained below.
\begin{itemize}
    \item The semantic distance $d_o(x,y)$ is defined as the distance between the semantic types of $x$ and $y$ in the ontology. Several approaches, e.g., \cite{Lee08co,Thiagarajan08cs,Pesquita09ss}, can be used to measure the semantic distance given the taxonomy in an ontology. We require the distance to be very large if the two semantic types are disjoint (i.e., having no common instance).
    \item The temporal distance $d_t(x,y)$ is defined as the time difference between the temporal annotations of $x$ and $y$ if both have temporal annotations.
    \item The spatial distance $d_s(x,y)$ is defined as the Euclidean distance between the spatial annotations of $x$ and $y$ if both have spatial annotations.
\end{itemize}
\end{itemize}
Currently we set the values of all the constants in the definition heuristically, but it is possible to optimize them by learning from annotated training data (e.g., \cite{Dodge2012dv}).

\subsection{Joint inference algorithm}\label{sec:joint:alg}
We start with the simple union of the video and text parse graphs and then iteratively apply three types of operators that make changes to the joint parse graph.
\begin{itemize}
    \item \textbf{Matching}. We match a node $a$ from the video parse graph with a node $b$ from the text parse graph, and merge them into a single node $c$ which inherits all the edges attached to either node $a$ or node $b$. To reduce the search space, we only match nodes that have a small distance between them (as defined in Eq.\ref{eq:dist}). The semantic type of the new node $c$ is set to be the more special one of the semantic types of node $a$ and node $b$. For example, if node $a$ is of semantic type ``Vehicle'' and node $b$ is of semantic type ``Food\_Truck'', then the semantic type of node $c$ is set to ``Food\_Truck'' because ``Food\_Truck'' is a sub-type of ``Vehicle''. The temporal and spatial annotations of nodes $a$ and $b$ are averaged before being assigned to the new node $c$. After the merging, if there exist two edges connecting to node $c$ that have the same direction and a small distance (as defined in Equation \ref{eq:dist}), then the two edges are also merged.
    \item \textbf{Deduction.} We insert a new subgraph into the joint parse graph, such that the prior probability of the joint parse graph as defined by the S/T/C-AOG is increased. Specifically, for each entity in the joint parse graph, we find the type of this entity in the S/T/C-AOG and insert an instantiation of its immediate surrounding subgraph in the AOG into the joint parse graph. Because the energy defined in section \ref{sec:joint:lh} actually penalizes the size of the joint parse graph, we require the insertion to be minimal, i.e., each inserted node or edge does not duplicate any existing node or edge. This can be achieved by following the deduction operator with a series of matching operators to match and merge the inserted nodes and edges with the existing parse graph.
    \item \textbf{Revision.} We either remove a subgraph from the parse graph, or change the annotation of a node or edge in the parse graph, in order to solve a conflict as defined in the S/T/C-AOG (e.g., an instantiation of zero probability, or instantiation of more than one child of an Or-node).
    Because the energy defined in section \ref{sec:joint:lh} penalizes such revisions, we require the removal or change to be minimal, i.e., no extra node or edge is removed or changed that is unnecessary to solve the conflict.
\end{itemize}

Our objective function is the joint posterior probability defined in Eq.\ref{eq:obj}. Because both the prior probability $P(pg_{jnt})$ (Eq.\ref{eq:prior}) and the conditional probability $P(pg_{vid},pg_{txt} | pg_{jnt})$ (Eq.\ref{eq:lh}) can be factorized by the entities and relations in the parse graph, we can compute the change to the joint posterior probability caused by each operator.

For the surveillance data that we focus on in this paper, a typical parse graph contains a few tens of nodes and edges, so we use depth-first search with pruning to optimize the joint parse graph.
For data of larger scales, it is straightforward to extend our algorithm with beam search.
Given the initial joint parse graph made of a simple union of the video and text parse graphs, we first apply the matching operators until no further matching is possible, then apply the revision operators to solve conflicts, and finally apply the deduction operators (with the follow-up matching operators) to deduce implicit information. Each operator typically contains a number of possibilities (e.g., multiple nodes in the text parse graph can be matched with a specific node in the video parse graph), and we exam all the possibilities in a depth-first manner. In order to reduce the search space, we prune the possibilities that have significantly higher energy than the others.

\textbf{Stop criterion for deduction.}
When applying the deduction operator, in some cases we may deduce multiple candidate subgraphs that have similar energy and are mutually exclusive (e.g., different children of an Or-node). In other words, there is significant uncertainty in the deduction as measured by information entropy. For example, we may observe a human in the video but cannot see what he is doing because of low resolution or occlusion, and based on the background knowledge we may infer a number of actions of the human that are equally possible. Since we adopt the open world assumption, it is reasonable to keep agnostic here and not add any new information. Specifically, we cancel the deduction operator if the entropy of the deduced subgraph is higher than a threshold:
\begin{eqnarray*}
	H(pg_{de}|pg_{jnt}) &=& -\sum_{i=1}^N P(pg_{de}^i|pg_{jnt}) \log P(pg_{de}^i|pg_{jnt}) \\
	&>& \frac{\log N}{c}
\end{eqnarray*}
where $pg_{jnt}$ is the joint parse graph before applying a deduction operator, $pg_{de}$ denotes the subgraph inserted by the deduction operator, $N$ is the number of possible candidate subgraphs that can be deduced by the operator, and $c>1$ is a pre-specified constant.

Note that in video parsing we produce multiple candidate video parse graphs. So for each video parse graph we run the joint inference algorithm to find a joint parse graph, and then we output the parse graph triple $\langle pg_{vid}, pg_{txt}, pg_{jnt} \rangle$ with the highest joint posterior probability.
Although the algorithm described here only outputs a single optimal joint parse graph for each pair of input video and text parse graphs, it is easy to adapt the algorithm to output multiple candidate joint parse graphs along with their joint posterior probabilities.


\section{Answering natural language queries}\label{sec:query}
The joint parse graph is a semantic representation of the objects, scenes and events contained in the input video and text, which is very useful in many applications.
In this section, we show how the joint parse graph can be used in semantic queries to answer questions, such as (a) questions in the forms of who, what, when, where and why; and (b) summary of scenes or events, e.g. how many persons.

\subsection{Text query parsing}
Given a simple plain text question about the objects, scenes and events in the input video and text, we parse the question into a formal semantic query representation. Since our joint parse graph can be represented in RDF, we represent a query using SPARQL, the standard query language for RDF \cite{SPARQL}.

The steps of text query parsing are identical to those in the text parsing module introduced in section \ref{sec:text}. The attribute grammar for analyzing the dependencies is extended to include interrogative wh-words such as \emph{who}, \emph{what}, \emph{when}, \emph{where} and \emph{why}. These wh-words indicate what the objective of the query is, and the rest of the text query defines the query conditions. For example, the question ``Who has a cap?'' is parsed into the following dependencies and attributes:
\begin{center}\renewcommand{\arraystretch}{1}
\begin{tabular}{l}
nsubj(Event, who)\\
root(ROOT-0, Event); Event.type = Possess\\
det(Object, Det)\\
dobj(Event, Object); Object.type = Cap
\end{tabular}
\end{center}

The first dependency indicates that the query is asking for the agent of an event. The rest of the dependencies specify what the event type (``Possess'') and the event patient (``Cap'') are. These query conditions are then converted into the SPARQL format:
\begin{verbatim}
    SELECT ?name ?who1
    WHERE {
        ?who1 hasName ?name.
        ?cap4 rdf:type Cap.
        ?has2 rdf:type Possess.
        ?has2 hasAgent ?who1.
        ?has2 hasPatient ?cap4.
    }
\end{verbatim}

Queries that require aggregates can also be expressed in SPARQL via 
aggregate functions (COUNT, MAX, etc.) and grouping. For instance, the question ``how many persons are there in the scene'' is parsed into the following SPARQL query:
\begin{verbatim}
 SELECT (COUNT(DISTINCT ?agent) as ?count)
 WHERE {
     ?agent rdf:type Human
 }
\end{verbatim}

After the SPARQL query is generated, it is evaluated by a SPARQL query engine which essentially performs graph matching of the patterns in the query condition with the joint parse graph stored in the RDF knowledge base. One implementation of the SPARQL query engine is included in Apache Jena \cite{Jena}. The values obtained from the query engine are then processed to produce the answers to the query.

\subsection{Computing answer probabilities from multiple interpretations}
\begin{figure}
\centering
\includegraphics[width=\columnwidth]{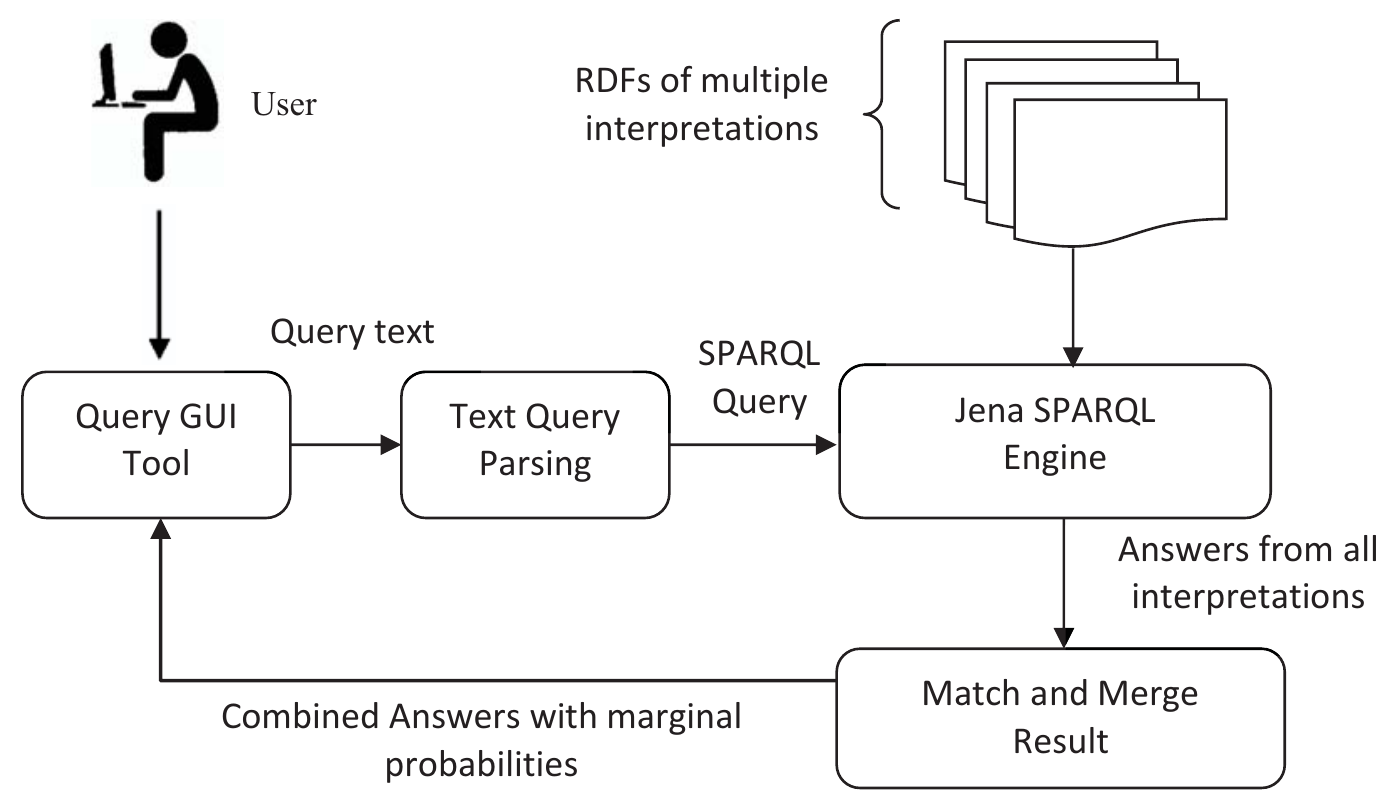}
\caption{The schematic diagram of querying multiple interpretations.}
\label{fig:prob_query}
\end{figure}

Our system can produce multiple joint parse graphs, each associated with its posterior probability. These joint parse graphs correspond to different interpretations of the input video and text. To answer a query accurately, we can execute the query on all the interpretations. The collected answers are then compared. We associate the answers from different interpretations by matching their semantic types and spatial-temporal annotations. 
For answers that are matched, their probabilities are combined. 
Formally, the probability of an answer $a$ is computed as:
\[
	P(a) = \sum_{pg} P(pg) \mathbbm{1}(pg \models a)
\]
where $P(pg)$ denotes the posterior probability of the joint parse graph $pg$ and $\mathbbm{1}(pg \models a)$ is the indicator function of whether parse graph $pg$ entails the answer $a$.
In this way, different possible answers can be ranked based on their marginal probabilities. A schematic diagram of the process is shown in Figure \ref{fig:prob_query}.

\subsection{Text query GUI tool}
\begin{figure}
\centering
\includegraphics[width=\columnwidth]{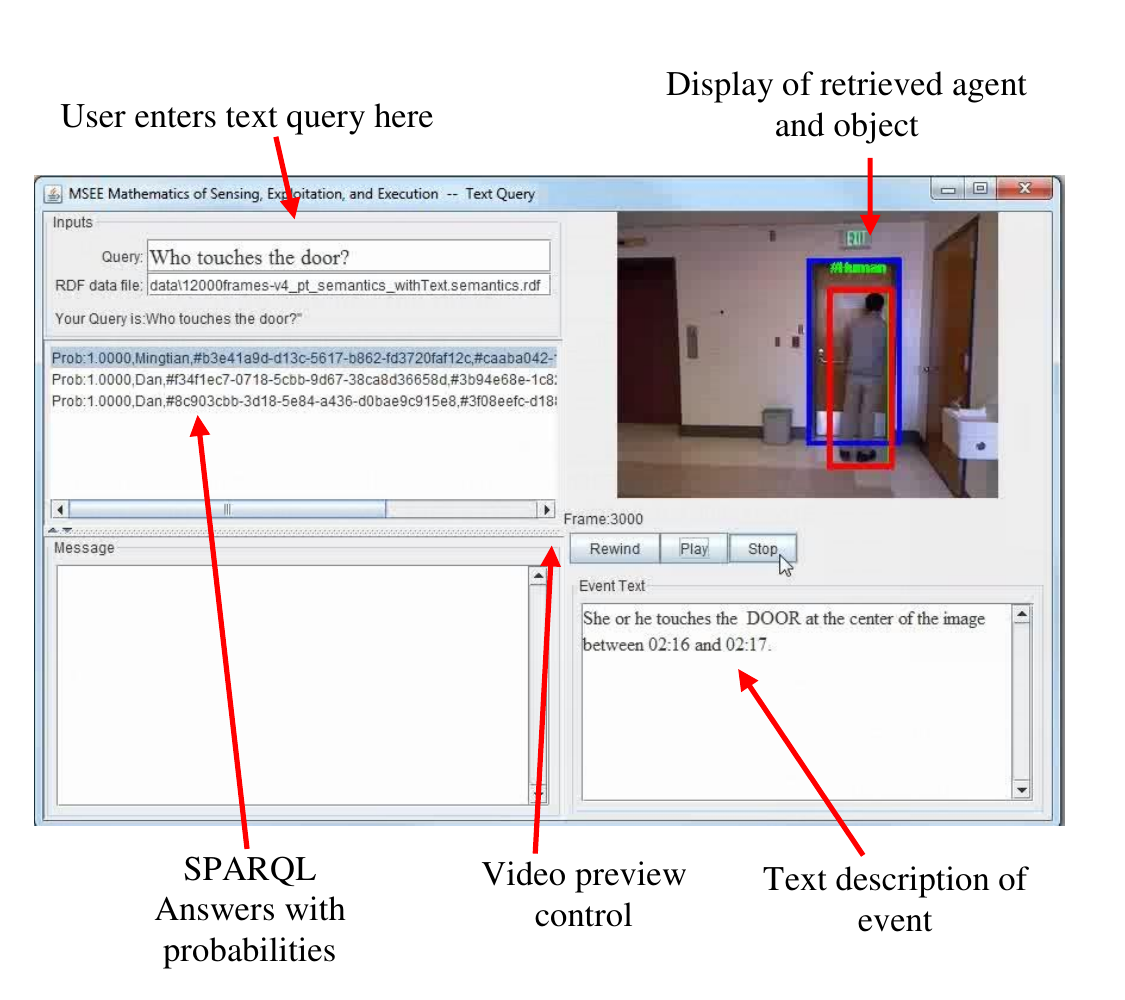}
\caption{Text Query GUI. See \url{http://www.stat.ucla.edu/~tukw/JointParsing/demo.html\#demo2} for a demonstration video.}
\label{fig:query_gui}
\end{figure}

We developed a GUI tool for text query answering. A screenshot of this tool is shown in Figure \ref{fig:query_gui}. With this tool, the user types a query in natural language text and the system parses the query and retrieves the result, which is shown on the left of the GUI. If the answer is an event, the text description of the event is also presented in the bottom right of the GUI (the text descriptions are generated automatically from the joint parse graphs, which will be discussed in section \ref{sec:exp:text_gen}).
The corresponding video segment of the retrieved answer is shown on the top-right with overlays surrounding the agent and patient of the event. With this tool, the user can efficiently query about the events occurring in the video and review the relevant video segments. In our experiment, each query response typically took a few seconds.

\section{Experiments}\label{sec:exp}
\subsection{Datasets}
We collected two datasets of surveillance videos and text descriptions in indoor (hallway) and outdoor (courtyard) scenes. Compared with other types of datasets (such as movies with screenplays and news report footages with closed captions), surveillance videos contain a well-defined set of objects, scenes and events, which is amenable to controlled experiments; the text accompanying surveillance videos provides descriptions of the scenes shown in the video, while other types of datasets may contain text that are not descriptive of the scenes such as dialogs and comments; in addition, on surveillance videos we can control the level of details in the text descriptions and study its impact on the performance of joint parsing.

\begin{figure}\centering
  \includegraphics[width=\columnwidth]{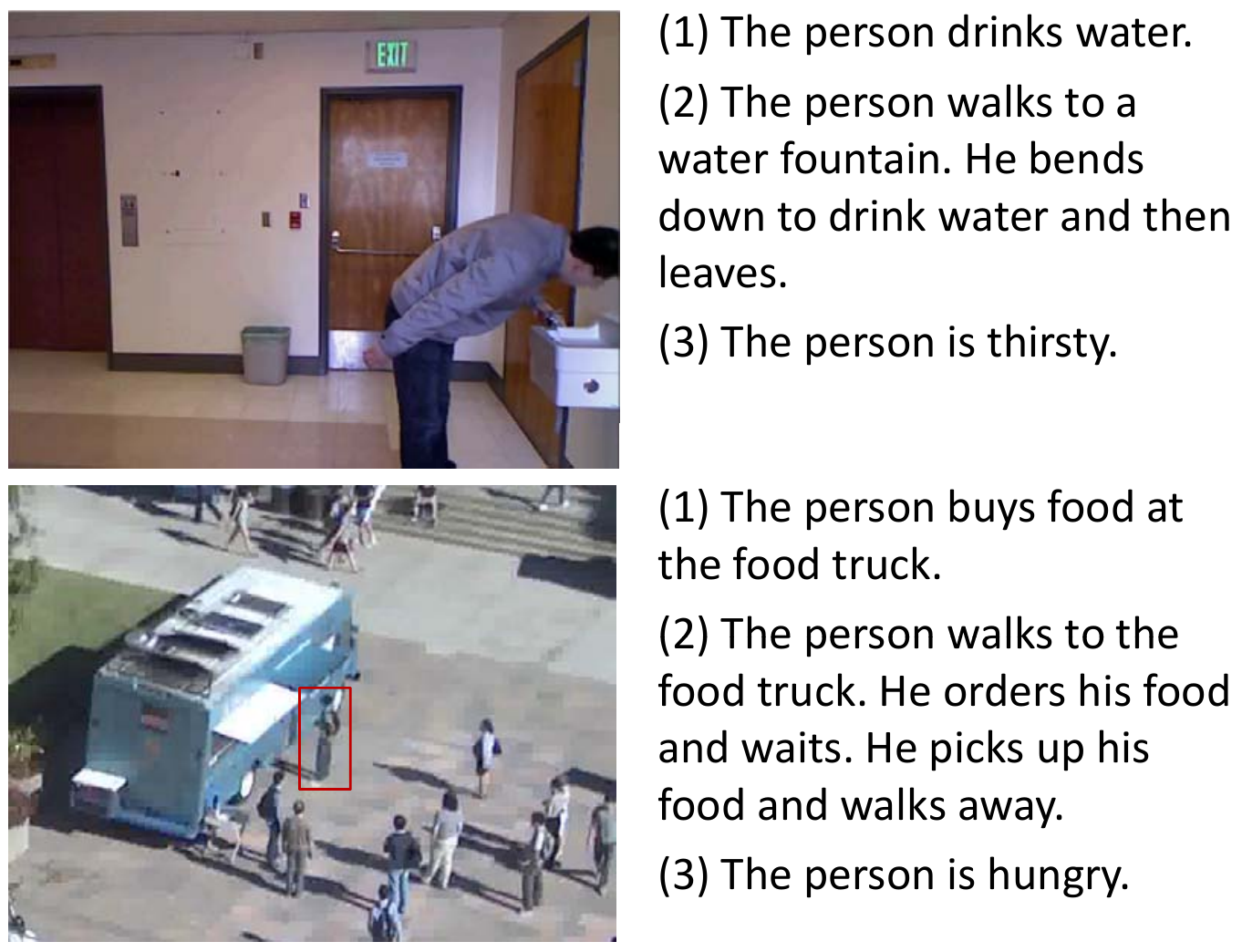}\\
  \caption{Examples from the hallway (top) and courtyard (bottom) datasets. The text descriptions shown on the right side are divided into three levels of details.}\label{fig:dataset}
\end{figure}

For each of the two datasets, we first recorded the video of the scene and divided the video into fifteen clips. Each video clip contains one or more events and lasts from a few seconds to over one minute. 
Based on the objects and events captured in the video, we manually constructed the ontology and the S/T/C-AOG for each dataset.
We then asked five human subjects from different countries and background to provide text descriptions by watching the video clips. 
The human subjects were asked to only describe the events and the involved agents, objects and fluents in the scenes using plain English. The ontology was also provided to the human subjects to restrict the scope of their description.
The human subjects were instructed to provide descriptions at three different levels of details, so we can study how the amount of text descriptions and their degrees of overlap with the video content can impact the performance of joint parsing. At the first level, the human subjects were told to give the simplest descriptions consisting of one or two sentences; at the second level, they were asked to give more details (typically less than ten sentences); at the third level, they were asked to provide information that is not directly observable in the video but can be inferred (typically zero to two sentences).
Figure \ref{fig:dataset} shows two examples from our datasets.

\subsection{Qualitative results}
\begin{table*}\centering
  \includegraphics[scale=0.6]{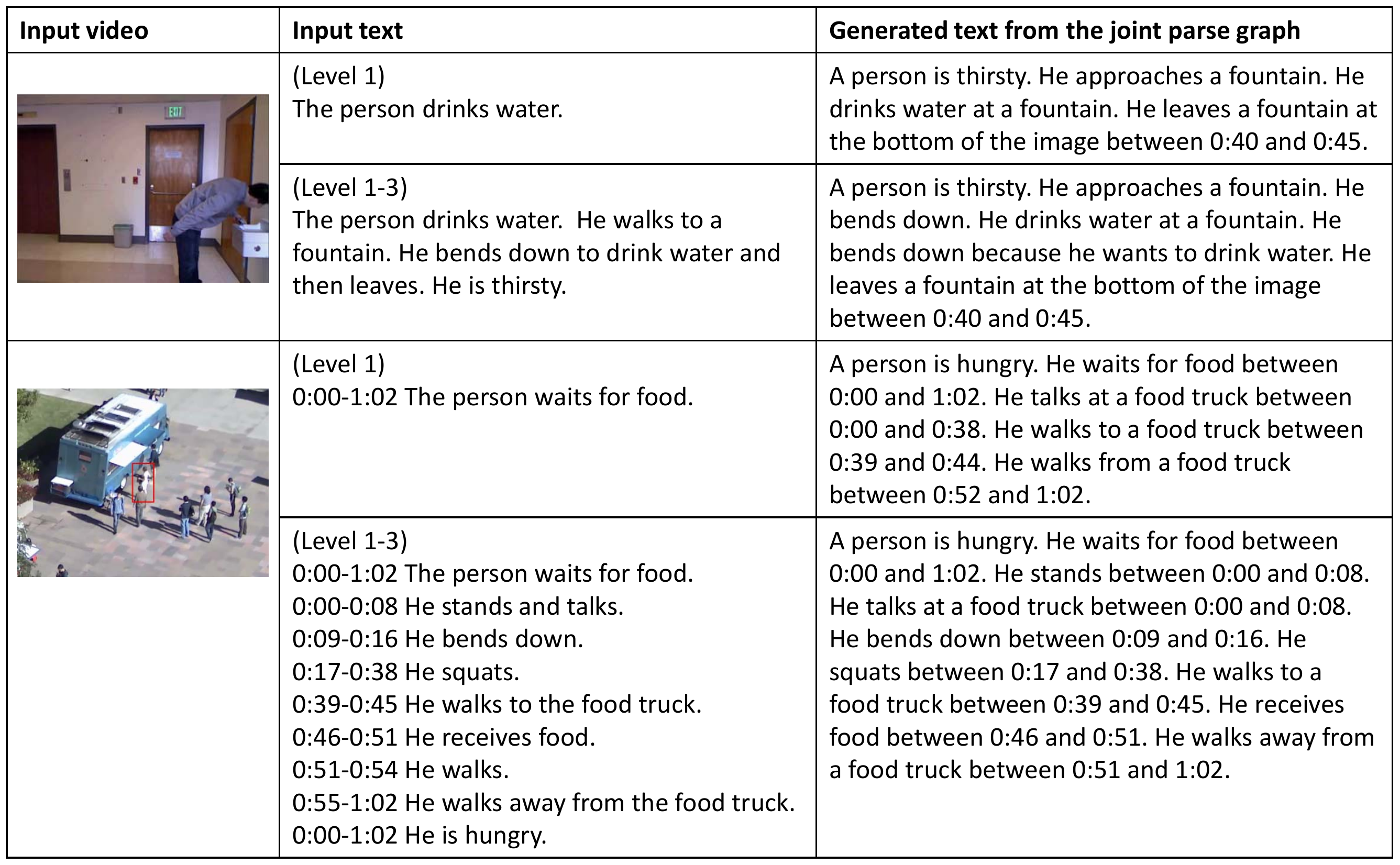}\\
  \caption{Examples of generated text from the joint parse graphs. For each input video, we show the results produced from joint parsing using level 1 text and using text of all three levels.}\label{tab:text_gen}
\end{table*}

We ran our system to produce video parses, text parses and joint parses from the datasets.
On a typical video-text pair, video parsing takes a few minutes to finish when running on a 12-node computer cluster, while text parsing and joint parsing finish within a few seconds on a desktop computer.
We compared the information contained in the three types of parse graphs and made the following observations.
\begin{itemize}
\item Prominent objects and events are typically contained in both the video and the text parse graphs, e.g., a person drinking at the water fountain.
\item Low-level visible details of objects, events and fluents are often contained in the video parse graphs but not in the text parse graphs, e.g., a person approaching the water fountain before drinking and the door closing after a person enters.
\item Two types of information are typically contained in the text parse graphs but not in the video parse graphs. 
The first is information invisible in the video but can be inferred, e.g., a person is thirsty (because he is drinking water). 
The second type of information is visible in the video but is not correctly detected by the video parser.
\item The joint parse graphs may contain information that can be found in neither the video nor the text parse graphs, e.g., low-level details of objects and events that are not detected by the video parser, or invisible information that is not mentioned in the input text. Such information is deduced in joint inference from other information in the video and text parse graphs.
\end{itemize}

\subsubsection{Text generation}\label{sec:exp:text_gen}
We extend the text generator described in Yao et al. \cite{Yao10i2t} to generate text descriptions from our joint parse graphs. The generated text presents the content of parse graphs in a user-friendly manner and helps non-expert users to understand the results from joint parsing.

The text generation process goes through two modules: sentence planner and sentence realizer. The sentence planner selects the content to be expressed, specifies hard sentence boundary, and provides information about the content. During this process, the information from the parse graph is converted to a functional description \cite{Langkilde98gt}, which specifies elements of a sentence such as the event, subjects, objects, and their functional properties. With a functional description as the input, the sentence realizer generates the sentence by determining a grammatical form and performing word substitution. A head-driven phrase structure grammar (HPSG) \cite{Pollard94hd} of English syntax is used to generate text sentences.

Table \ref{tab:text_gen} shows some examples of the generated text.
By comparing the input text and the generated text, it can be seen that when the input text is simple, the majority of the information in the joint parse graph comes from video parsing as well as deduction in joint inference; with more input text, an increasing part of the joint parse graph comes from text parsing.

\begin{figure*}\centering
	\subfigure[Hallway]{\includegraphics[scale=.6]{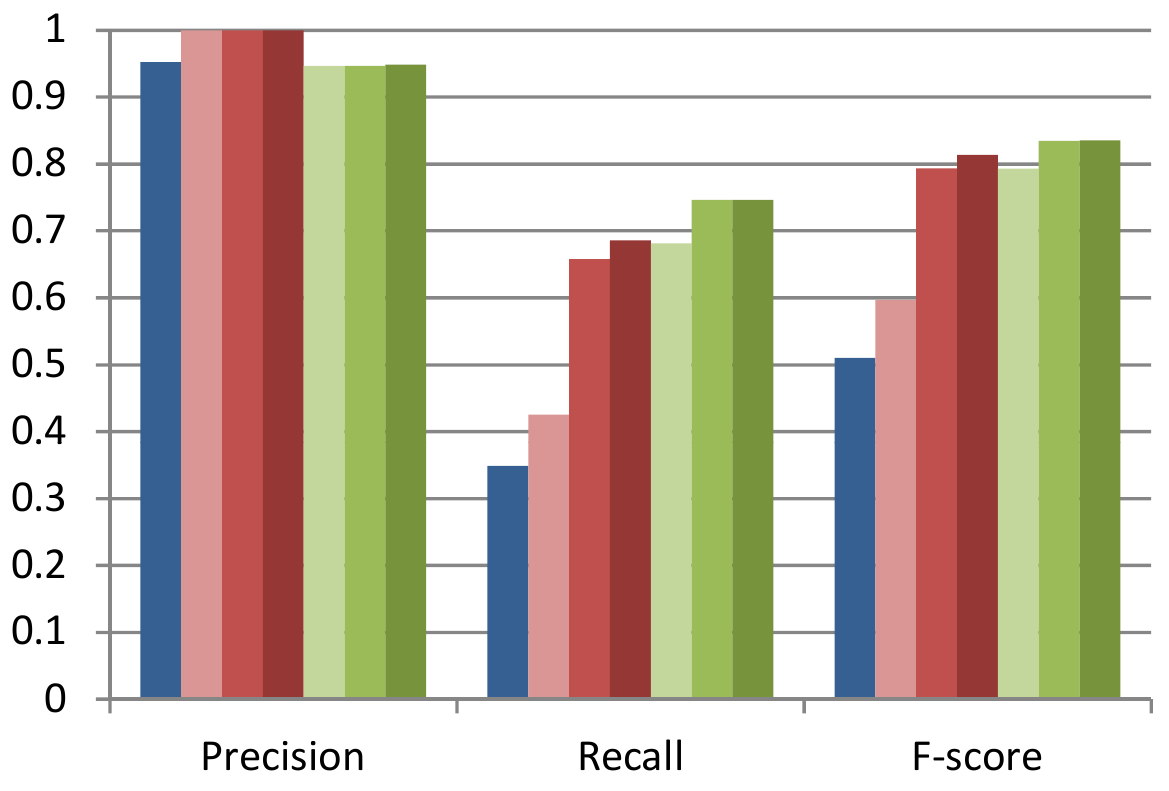}}
	\subfigure[Courtyard]{\includegraphics[scale=.6]{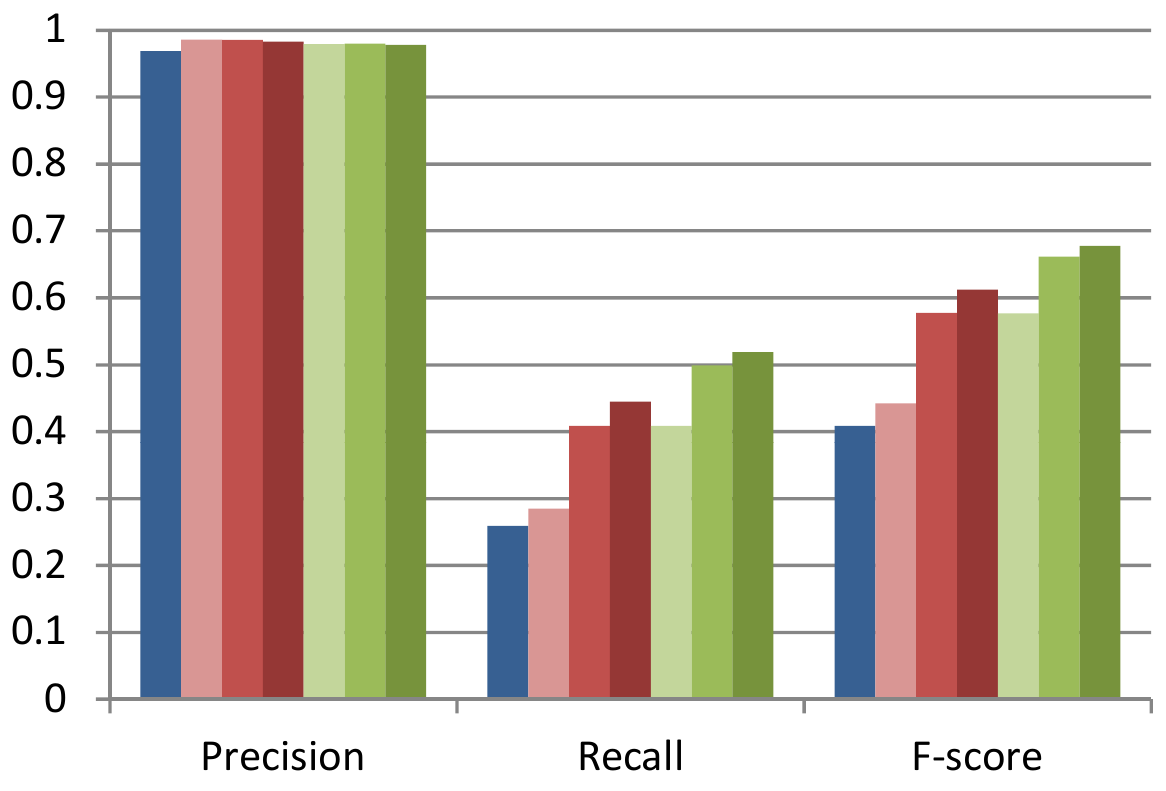}}
	\includegraphics[scale=.6]{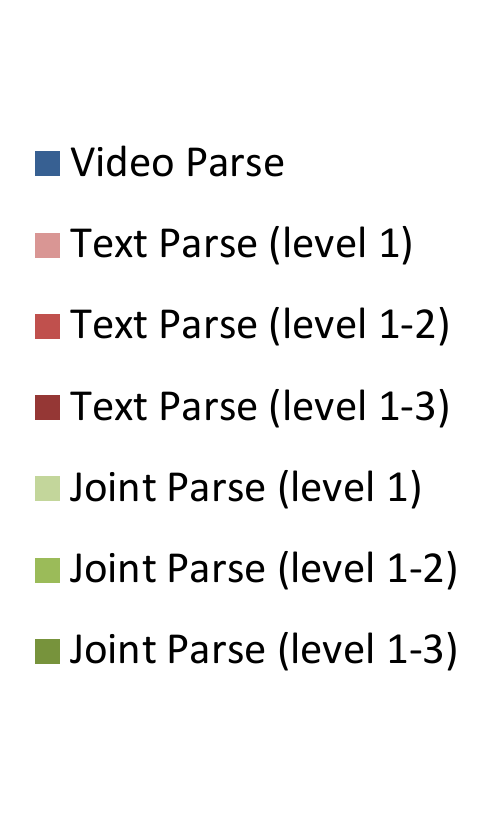}\\
	\caption{The average precision, recall and F-score of the video, text and joint parse graphs in comparison against the ground truth parse graphs constructed from merged human text descriptions.}\label{fig:exp:pr}
\end{figure*}

We also evaluated the generated text based on the joint parse graph against the merged text descriptions from human subjects, using quantitative metrics such as BLEU \cite{Papineni2002bleu} and METEOR \cite{Banerjee2005meteor} which were developed for evaluating machine translation. We find the resulting scores uninformative about the quality of joint parsing, because 1) the automatic text generator often chooses different words and expressions than human (e.g., ``bend because he wants to drink'' vs. ``bend to drink''); 2) the generated text often contains information that is not included in the human text descriptions, such as the information from video parsing and deduction (e.g., low-level visual details). These differences are heavily penalized by the machine translation evaluation metrics, although they either are unimportant or should be rewarded for evaluating joint parse graphs.
Therefore, we adopt two alternative quantitative evaluation approaches as introduced in the next subsection.

\subsection{Quantitative evaluation}
On each video clip in the datasets, for each human subject we used the text descriptions of the video clip as the text input to produce three joint parse graphs. The first joint parse graph is based on the level-1 text descriptions; the second is based on the combined text from level 1 and 2; and the third is based on the text of all three levels.
We consider four baseline parse graphs: the video parse graph and the three text parse graphs based on level 1 text, level 1+2 text and text of all three levels.
We evaluated the joint parse graphs as well as the baseline parse graphs using two different evaluation approaches.


\subsubsection{Comparison against ground truth}
\textbf{Ground truth}. 
For each video clip, we constructed the ground truth parse graph by merging the parses of all the text descriptions from all the human subjects except the one whose text description was used as the input in the current run of experiments. The merged parse graph was manually checked to ensure correctness. 

\textbf{Precision and recall}.
We compared the parse graphs to be evaluated against the ground truth parse graph and measured the precision, recall and F-score. Let $pg$ represent the parse graph to be evaluated, $pg^*$ the ground truth parse graph, $pg \cap pg^*$ the overlap between the two parse graphs, and $\|pg\|$ the size of parse graph $pg$.
\begin{eqnarray*}
\mathrm{Precision} &=& \frac{\|pg \cap pg^*\|}{\|pg\|} \\
\mathrm{Recall} &=& \frac{\|pg \cap pg^*\|}{\|pg^*\|}
\end{eqnarray*}
F-score is defined as the harmonic mean of precision and recall. 
In computing precision and recall, in order to estimate the overlap between the parse graph to be evaluated and the ground truth parse graph, we ran depth-first search using the matching operator described in section \ref{sec:joint:alg} to match the nodes and edges in the two parse graphs.
Since there might be information that is correct but missing from our ground truth parse graph constructed from merged human text descriptions, we also manually checked the parse graph to be evaluated when computing its precision.

Figure \ref{fig:exp:pr} shows the precision, recall and F-score of the baseline parse graphs and joint parse graphs averaged over all the video clips and human subjects of each dataset.
It can be seen that all the precision scores are close to 1, suggesting the correctness of the information contained in all types of parse graphs. On the other hand, the recall and F-score of different types of parse graphs vary significantly. Each level of joint parse graphs has higher recall and hence higher F-score than the video parse graphs and the text parse graphs of the same level, which demonstrates the advantage of joint parsing. In particular, the joint parse graphs based on level-1 text have significantly higher recall and F-score than both the video parse graphs and the level-1 text parse graphs, implying that the information from the video and the level-1 text is highly complementary; it also shows that providing a simple text description (i.e., level-1 text) to a video can already boost the performance of video understanding to the level close to a full description of the video (i.e., all three levels of text combined). Adding more text (level 2 and 3) into joint parsing can be seen to further improve the joint parse graphs but with diminishing return.

\begin{figure*}\centering
	\subfigure[Hallway]{\includegraphics[scale=.6]{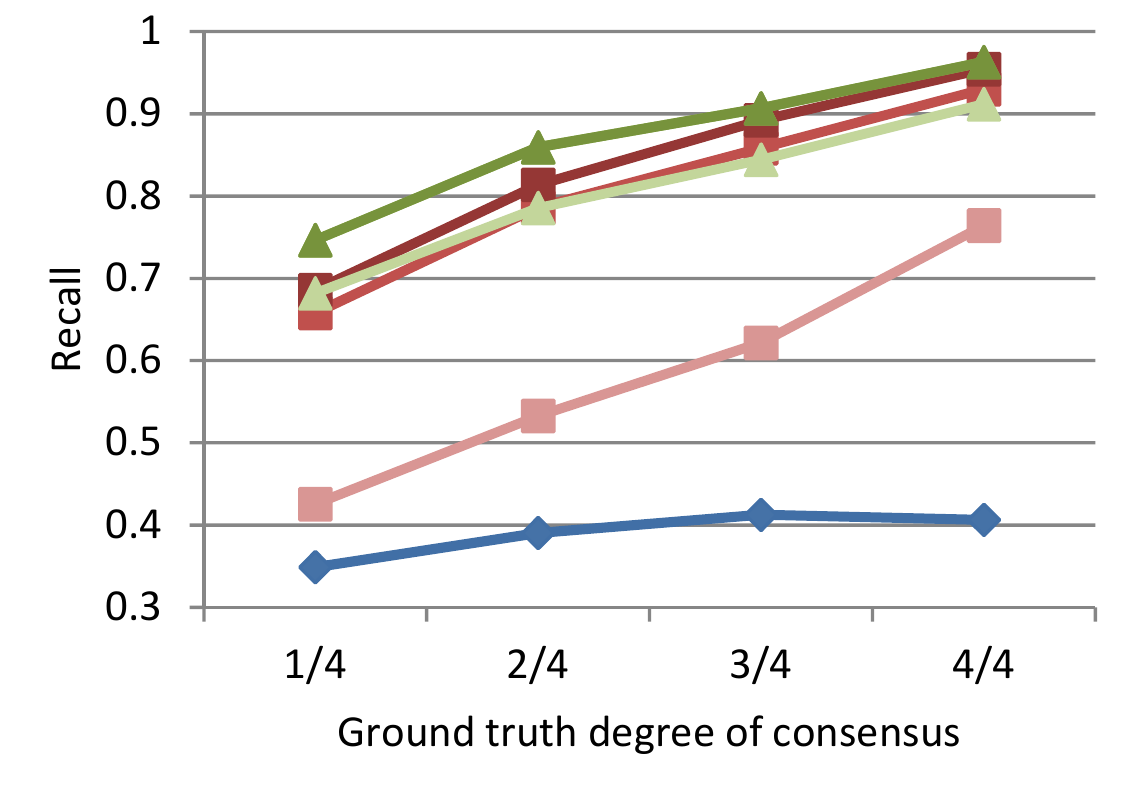}}
	\subfigure[Courtyard]{\includegraphics[scale=.6]{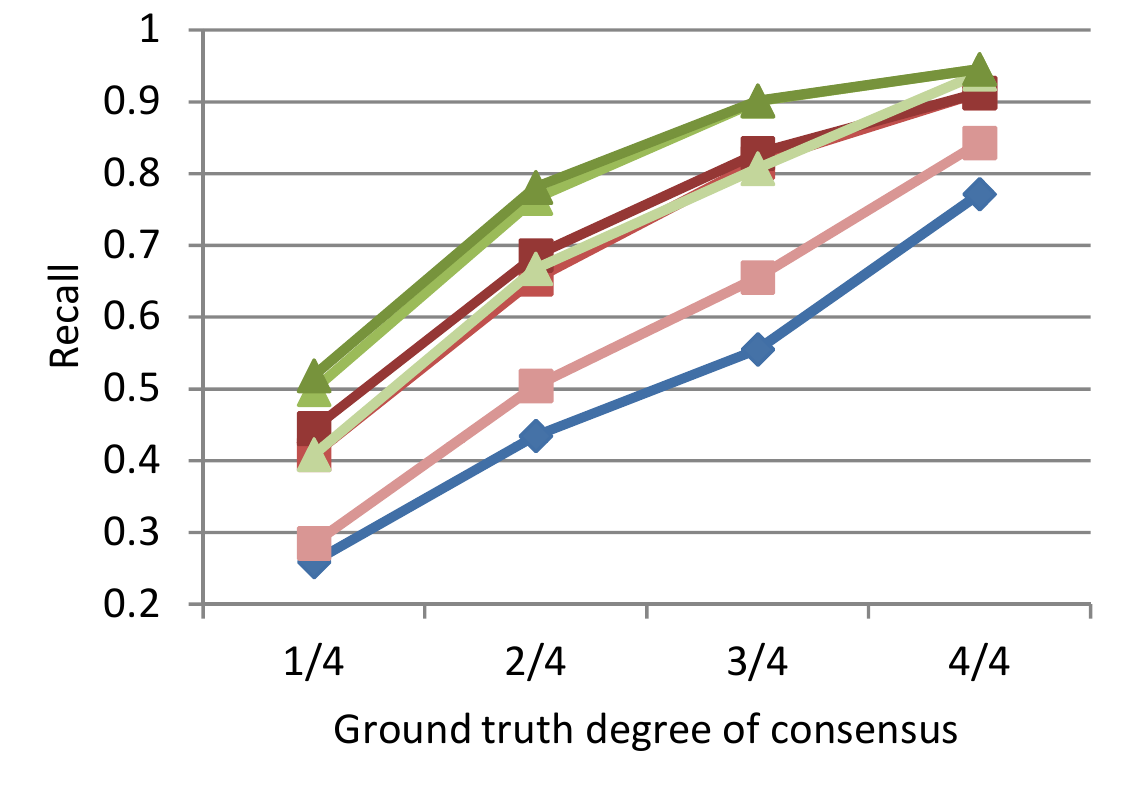}}
	\raisebox{2ex}{\includegraphics[scale=.6]{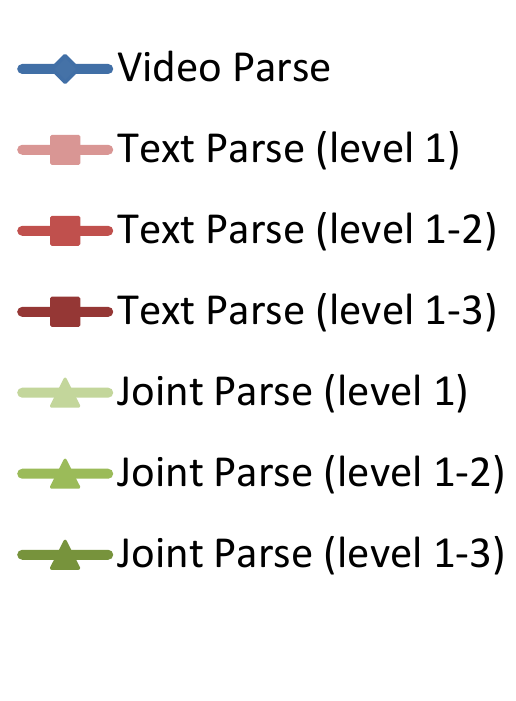}}\\
	\caption{The average recall of the video, text and joint parse graphs evaluated against the ground truth parse graphs re-constructed based on different degrees of consensus.}\label{fig:exp:r}
\end{figure*}

Comparing the results of the two datasets, we can see that the recall scores of the text and joint parse graphs on the courtyard dataset are much lower than those on the hallway dataset. This is because the courtyard dataset contains more complicated events and therefore the text descriptions from the five human subjects are often more different. Below is an example of such difference on a courtyard video clip:
\begin{enumerate}
\item The person walks.
\item The person walks and \emph{talks}.
\item The person walks \emph{out of a building}.
\item The person walks \emph{across the courtyard}.
\end{enumerate}
Since none of the text descriptions cover all the details of the event, the average recall of the text parse graphs is low. The joint parse graphs are produced with the text parse graphs as input, so their average recall is also lower on the courtyard dataset.

To compensate for this phenomenon in computing recall, we changed the ground truth parse graphs by keeping only the entities and relations that are mentioned by a minimal number of human subjects. The number of human subjects that mention an entity or relation indicates the degree of consensus among the human subjects regarding the existence and importance of the entity or relation. The maximal degree of consensus is $4/4$ in our experiments because we used the text from four of the five human subjects to construct the ground truth parse graphs (excluding the provider of the input text of joint parsing).
Figure \ref{fig:exp:r} shows the average recall scores of all types of parse graphs evaluated against the re-constructed ground truth parse graphs based on different degrees of consensus. 
It can be seen that all the recall scores are improved with the increase of the ground truth degree of consensus. Most of the improvements are very large, implying that the text descriptions from different human subjects are indeed quite different, leading to significant changes of the ground truth parse graphs at different degrees of consensus. At the highest degree of consensus, the recall scores of most of the parse graphs are over 0.9, suggesting that these parse graphs cover the most important entities and relations in the scene. Among all types of parse graphs, only the video parse graphs on the hallway dataset do not have large improvement of recall, suggesting the limitation of the video parser on the hallway dataset.

\begin{figure}\centering
	\includegraphics[width=\columnwidth]{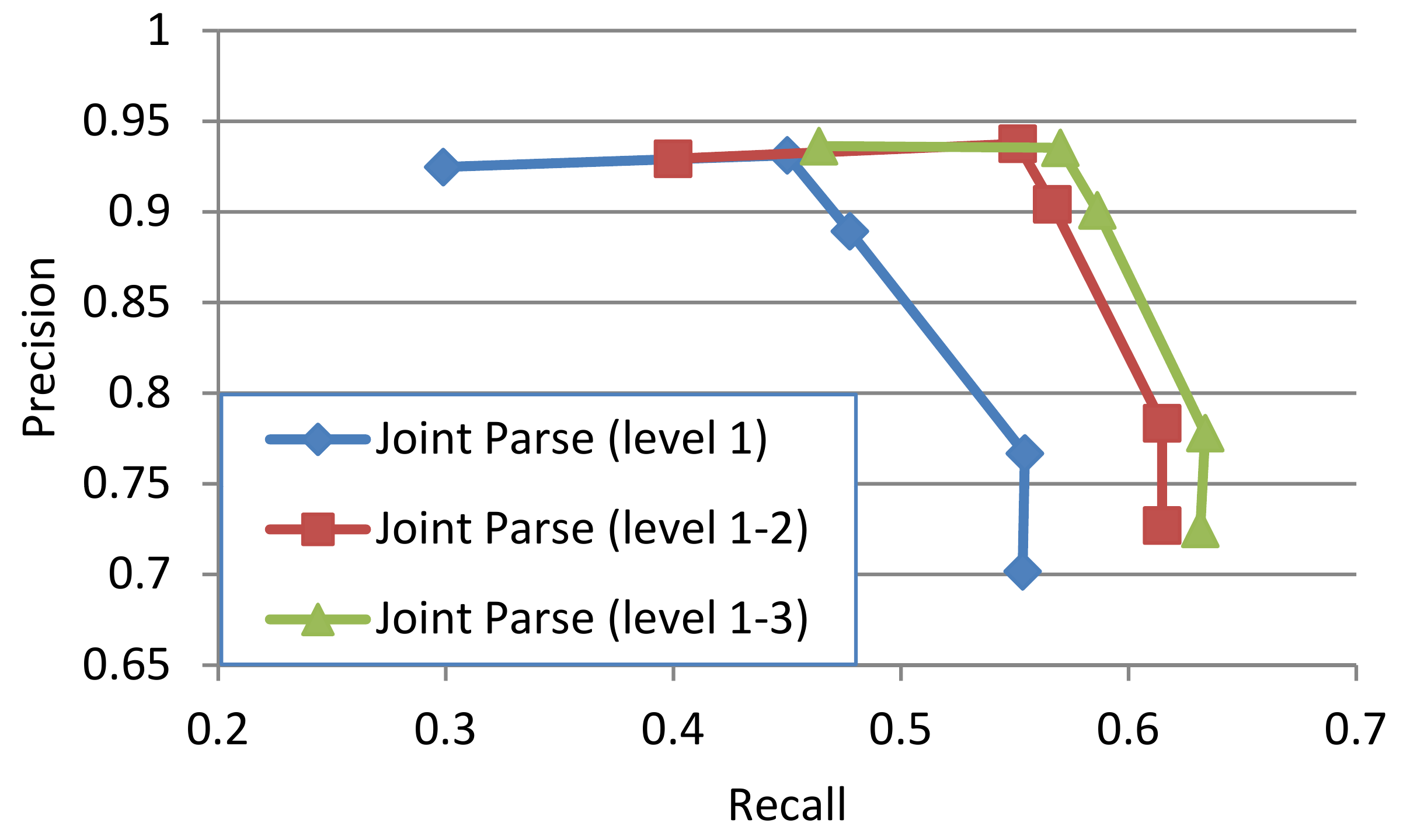}\\
	\caption{The influence of the degree of deduction on the precision and recall of joint parse graphs.}\label{fig:exp:deduction}
\end{figure}

\textbf{Degree of deduction}.
We further studied the influence of the degree of deduction on joint parsing. On a subset of the courtyard dataset that contain complex events, we adjusted the value of $\alpha_\phi$ to control the degree of deduction during joint inference and then measured the precision and recall of the resulting joint parse graphs. Figure \ref{fig:exp:deduction} shows the results. It can be seen that with an increasing degree of deduction, at first the recall is greatly improved without decreasing the precision, suggesting that the deduced information is mostly correct; however, with further deduction the recall is improved modestly while the precision drops dramatically, which is caused by the increase of erroneous information being deduced; eventually, the recall stops improving while the precision continues to drop, implying that all the additional information being deduced is false.

\subsubsection{Evaluation based on query answering}
\begin{figure*}\centering
	\subfigure[Hallway]{\includegraphics[scale=.6]{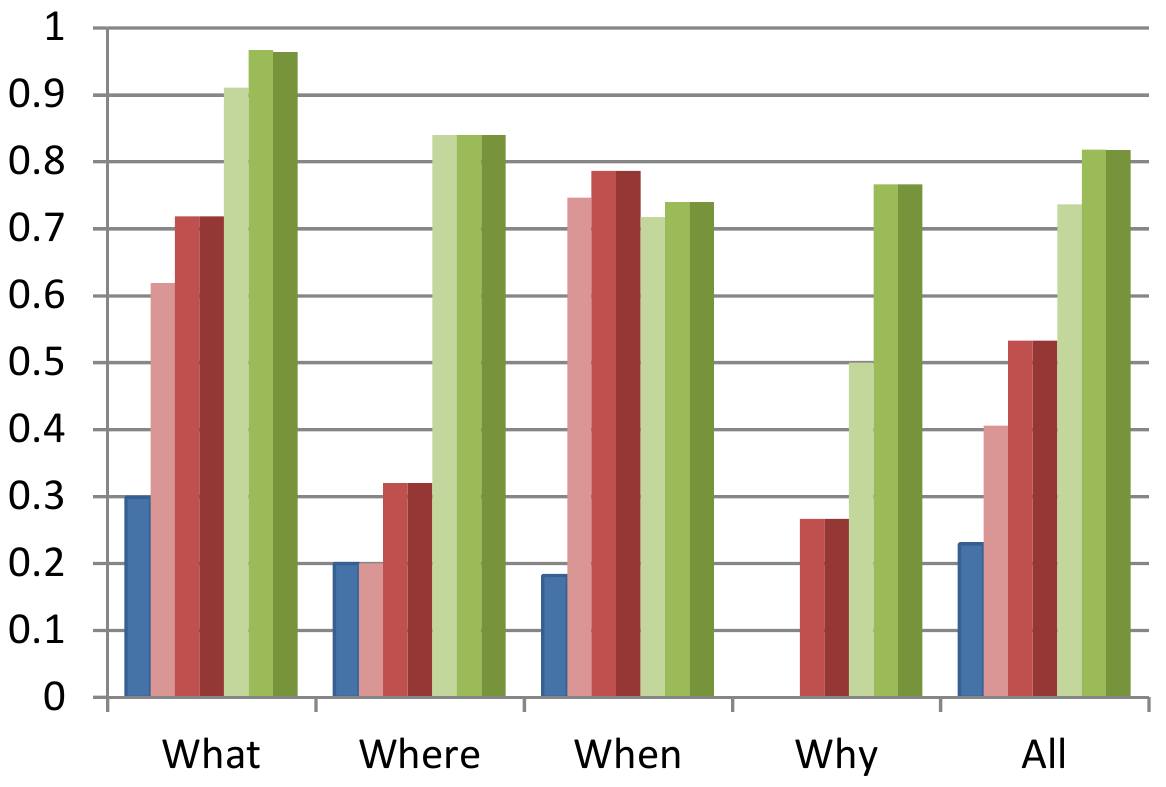}}
	\subfigure[Courtyard]{\includegraphics[scale=.6]{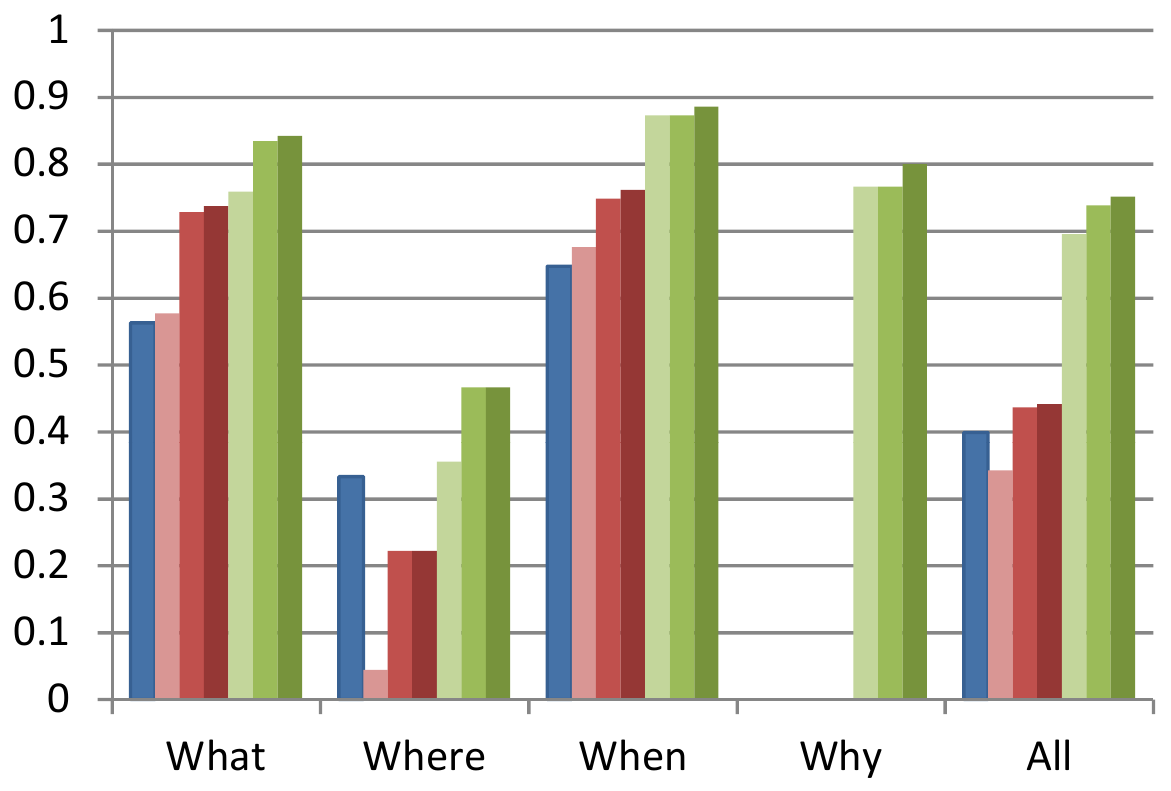}}
	\includegraphics[scale=.6]{exp_pr_legend}\\
	\caption{The average F-score of the what, where, when and why queries and of all the queries based on the video, text and joint parse graphs in the query answering experiments.}\label{fig:exp:query}
\end{figure*}

In section \ref{sec:query} we have introduced how the joint parse graph produced by our system can be used in query answering. Here we evaluate the quality of the joint parse graph by measuring the accuracy of query answering. This evaluation approach of video understanding is novel in that it directly measures the utility of the video understanding system in human-computer interaction, which goes beyond the conventional evaluation frameworks such as those based on classification or detection.

\begin{table}
\begin{tabular}{|p{0.25\columnwidth}|p{0.35\columnwidth}|p{0.25\columnwidth}|}
\hline
\textbf{Video clip} & \textbf{Queries} & \textbf{Correct answers} \\\hline
\multirow{5}{*}{\parbox{0.25\columnwidth}{\smallskip\includegraphics[width=0.25\columnwidth]{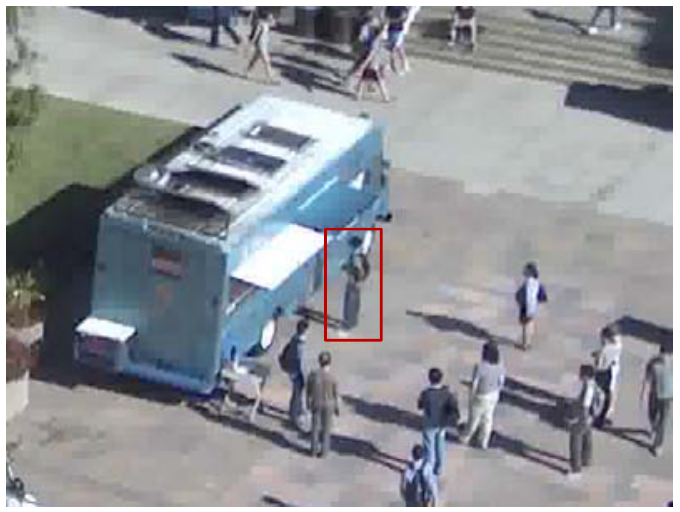}}} &
Who is there in the designated area? & A human. \\\cline{2-3}
& What is he doing? & Buying food; waiting; walking; picking up food. \\\cline{2-3}
& Where does he buy food? & At the food truck. \\\cline{2-3}
& Why does he buy food? & He is hungry. \\\cline{2-3}
& When does he pick up food? & 0:46-0:51. \\\hline
\end{tabular}
\smallskip
\caption{An example sequence of natural language queries and the correct answers.}\label{tab:exp:queries}
\end{table}

For each video clip in our datasets, we constructed a natural language query set that mimics the scenario in which a human user continuously queries the computer in order to acquire information of the objects, scene and events captured in the video clip. The natural language queries can take the forms of who, what, when, where and why. We also asked a human subject to provide the correct answers for each query set.
Table \ref{tab:exp:queries} shows an example query set and the correct answers.
We then entered the queries into our system and retrieved the answers.
Please see \url{http://www.stat.ucla.edu/~tukw/JointParsing/demo.html\#demo2} for a demonstration video of a user continuously asking queries through our GUI tool.

We compared the answers produced by the system against the correct answers provided by human. For each who/what/where/why query, we manually matched the system answers with the human answers and computed the precision (the percentage of the system answers that are correct) and recall (the percentage of the human answers that are returned by the system). For each when query, we checked the overlap between the time range returned by the system and that provided by human, based on which we computed the precision (the percentage of the time range returned by the system that is correct) and recall (the percentage of the correct time range that is covered by the system answer). We then computed the F-score which is the harmonic mean of precision and recall.

Figure \ref{fig:exp:query} shows the average F-score of the what, where, when and why queries and of all the queries based on the video, text and joint parse graphs. The results of the who queries are not shown because the video clips used in the experiments either contain or designate a single person and therefore the who queries become trivial to answer. 
From the results we can see that query answering based on the joint parse graphs clearly outperforms that based on the video and text parse graphs. Note that in comparison with the experimental results from the first evaluation approach (Figure \ref{fig:exp:pr}), in the query answering experiments the joint parse graphs have a significantly larger advantage over the video and text parse graphs. This is because answering a query involves perfect matching of the conditions of the query with one or more subgraphs of the parse graph, and therefore any small error in the parse graph would be magnified in the evaluation of query answering.
Among the four types of queries shown in Figure \ref{fig:exp:query}, the joint parse graphs have the largest advantage on the why queries over the video and text parse graphs, because the causal relations required for answering the why queries are typically not detected in video parsing and not always mentioned in the input text but can be inferred by the deduction operator in joint inference. 
The advantage of the joint parse graphs on the where queries is also large, because in video parsing many events being queried are not detected, and in the text the location information is sometimes skipped (as can be seen in the examples in Figure \ref{fig:dataset}), while in joint parsing additional location information can be deduced.
For the when queries, the average F-score of the text parse graphs is well below 1.0 although the time annotations provided by human subjects in the input text are quite accurate, which is because some of the events being queried are not described in the input text and therefore the corresponding when queries cannot be answered based on the text parse graphs.

\section{Conclusion and Discussion}\label{sec:conc}
We propose a multimedia analysis framework for parsing video and text jointly for understanding events and answering queries. 
Our framework produces a parse graph of video and text that is based on the spatial-temporal-causal And-Or graph (S/T/C-AOG) and represents the compositional structures of spatial information (objects and scenes), temporal information (actions and events) and causal information (causalities between events and fluents).
We present a probabilistic generative model for joint parsing that captures the relations between the input video/text, their corresponding parse graphs and the joint parse graph. Based on the model we propose a joint parsing system that consists of three modules: video parsing, text parsing and joint inference. In the joint inference module, we produce the joint parse graph by performing matching, deduction and revision on the video and text parse graphs.
We show how the joint parse graph produced by our system can be used to answer natural language queries in the forms of who, what, when, where and why. The empirical evaluation of our system shows satisfactory results.

For future work,
first, we will further improve both video parsing and text parsing with respect to parsing quality and efficiency. 
Second, currently we manually construct the S/T/C-AOG to model objects, scenes and events; in the future we want to investigate automatic approaches to learning AOG from data, which would facilitate the application of our approach to video-text data from novel domains.
Third, we plan to test our system on video and text data more challenging than surveillance videos with human intelligence descriptions, e.g., news report data, which contains more diversified contents and larger gap between video and text and requires more sophisticated background knowledge in joint parsing.

\section*{Acknowledgment}
The work is supported by the DARPA grant FA 8650-11-1-7149, the ONR grant N00014-10-1-0933 and N00014-11-C-0308, and the NSF CDI grant CNS 1028381.
We want to thank Mingtian Zhao, Yibiao Zhao, Ping Wei, Amy Morrow, Mohamed R. Amer, Dan Xie and Sinisa Todorovic for their help in automatic video parsing.

\bibliographystyle{ieeetran}
\bibliography{joint}

\end{document}